\documentclass{article}

\usepackage{arxiv}

\usepackage[utf8]{inputenc} 
\usepackage[T1]{fontenc}    
\usepackage{hyperref}       
\usepackage{url}            
\usepackage{booktabs}       
\usepackage{amsfonts}       
\usepackage{nicefrac}       
\usepackage{microtype}      
\usepackage{lipsum}
\usepackage{graphicx}
\usepackage{natbib}
\usepackage{tabularx}
\graphicspath{ {./images/} }

\usepackage[titletoc, title]{appendix}

\newcolumntype{C}{>{\centering\arraybackslash}X}

\usepackage{caption} 
\usepackage{tikz}
\usetikzlibrary{
    shapes.geometric,
    shapes.symbols,
    shapes.misc,
    arrows.meta,
    positioning
}
\tikzset{
    boxstyle/.style={
        rectangle, rounded corners,
        draw, align=center,
        minimum width=1.1cm,
        minimum height=0.5cm,
        fill=#1
    },
    arrowstyle/.style={
        ->, thick
    },
    cylinderstyle/.style={
        cylinder, shape border rotate=90,
        aspect=1, draw,
        minimum height=1.5cm,
        minimum width=0.5cm,
        fill=#1
    },
    diamondstyle/.style={
           shape=diamond,draw=blue!50!gray,
           ultra thick,fill=red!25!white,
           minimum height=3em,
           minimum width=4em
    }
}

\title{RoLegalGEC: Legal Domain Grammatical Error Detection and Correction Dataset for Romanian}

\author{
 Mircea Timpuriu\textsuperscript{1}, Mihaela-Claudia Cercel\textsuperscript{2,3}, and Dumitru-Clementin Cercel\textsuperscript{1,}\thanks{Corresponding author.} \\
  \textsuperscript{1}National University of Science and Technology POLITEHNICA Bucharest, Bucharest, Romania\\
   \textsuperscript{2}Paris 1 Panthéon-Sorbonne University, Paris, France \\
  \textsuperscript{3}University of Bucharest, Bucharest,
 Romania
  \\
\texttt{mirceatimpuriu@gmail.com,dumitru.cercel@upb.ro} }

\begin{document}
\maketitle
\begin{abstract}
The importance of clear and correct text in legal documents cannot be understated, and, consequently, a grammatical error correction tool meant to assist a professional in the law must have the ability to understand the possible errors in the context of a legal environment, correcting them accordingly, and implicitly needs to be trained in the same environment, using realistic legal data. However, the manually annotated data required by such a process is in short supply for languages such as Romanian, much less for a niche domain. The most common approach is the synthetic generation of parallel data; however, it requires a structured understanding of the Romanian grammar. In this paper, we introduce, to our knowledge, the first Romanian-language parallel dataset for the detection and correction of grammatical errors in the legal domain, RoLegalGEC, which aggregates 350,000 examples of errors in legal passages, along with error annotations. Moreover, we evaluate several neural network models that transform the dataset into a valuable tool for both detecting and correcting grammatical errors, including knowledge-distillation Transformers, sequence tagging architectures for detection, and a variety of pre-trained text-to-text Transformer models for correction. We consider that the set of models, together with the novel RoLegalGEC dataset,  will enrich the resource base for further research on Romanian.
\end{abstract}


\section{Introduction}
Addressing grammatical errors is one of the main concerns in the natural language processing (NLP) domain.
Thus, two specific NLP tasks have been defined, through specific benchmarks \citep{felice-briscoe-2015-towards} and shared tasks \citep{bryant-etal-2019-bea}, namely grammatical error detection (GED) and grammatical error correction (GEC). Due to a significant amount of resources, the research of error-related tasks for languages such as English is already very well established, through the creation of a well-defined error taxonomy and standardized evaluation mechanism \citep{bryant-etal-2017-automatic}, or methodologies of error classification and transformation of erroneous sequences into correct ones based on error nature \citep{omelianchuk2020gectorgrammaticalerror}. In the meantime, interest has been expressed for researching error detection and correction among low-resource languages as well, where the choices of smaller but better optimized models meant to fit the available resources better have proven very efficient for languages such as Indic \citep{sharma-bhattacharyya-2025-indigec}, Turkish \citep{kara2023gecturkgrammaticalerrorcorrection}, Ukrainian \citep{bondarenko-etal-2023-comparative}, and even Romanian \citep{cotet, info16030242}. Even considering these developments, the GEC and GED models still require substantial amounts of annotated data, which are difficult to obtain for low-resource languages. As such, NLP research for these languages has provided algorithms for synthetically generating erroneous sequences from originally correct sequences, such as backward application of language-specific grammatical rules \citep{xu-etal-2019-erroneous, bondarenko-etal-2023-comparative}, introduction of controlled noise into text \citep{kiyono2019empiricalstudyincorporatingpseudo}, and a large language model (LLM) prompting \citep{stahlberg-kumar-2024-synthetic, park2024chatlang8llmbasedsyntheticdata}.

The aforementioned fact that erroneous texts provide inferior value is most evident in fields such as the practice of law. Because the legal field is a space where nuance is subtle and words can be interpreted in several ways in the pursuit of essential conclusions, a text presented in this context must be grammatically and, more so, semantically correct. Interest in applying NLP to the English legal domain has grown, with techniques such as Transformer-based models \citep{vaswani2017attention, chalkidis-etal-2020-legal} and LLM-based methods \citep{dominguezolmedo2025lawmapowerspecializationlegal} used for the annotation and classification of legal corpora. Consequently, NLP research in the legal domain has extended to low-resource languages, such as Romanian, where the most notable contribution is LegalNERo \citep{pais-etal-2021-named, smuadu2022legal} for the named entity recognition task. However, to our knowledge, there are currently no Romanian-language legal datasets tailored specifically for tasks involving grammatical errors, for either GED or GEC. 

In our work, we aim to investigate the following main research questions:
\begin{itemize}
    \item RQ1: What are the advantages and disadvantages of few-shot prompting using common Romanian errors for specific errors of parts of speech?

     \item RQ2: What effect does the addition of GED error tags to the GEC input have on GEC model performance?

    \item RQ3: What are the advantages and disadvantages in terms of model performance in correlation with model size and pre-training language?

\end{itemize}

To address these questions, we first define an error taxonomy tailored to Romanian, based on a detailed analysis of common grammatical mistakes, yielding 20 distinct error types. Based on this taxonomy, we develop a methodology to synthetically generate each of the 20 error types by applying grammatical rules, introducing unintelligible noise into the text, or evaluating an LLM's ability to generate errors using common error examples. Ultimately, through our methodology, starting from comprehensive, powerfully annotated legal corpora, we propose the RoLegalGEC dataset, the first Romanian parallel dataset for the tasks of grammatical error detection and correction in the legal domain, which consists of 350,000 passages of legal and administrative documents that have been processed to render them erroneous, along with their grammatically correct counterparts, and an in-depth analysis of the placement and nature of errors in the incorrect sequence. Furthermore, using this dataset, we train and evaluate several architectures for both detection and correction tasks, such as knowledge distilled BERT \citep{sanh2020distilbertdistilledversionbert} and sequence tagging structures for error classification, along with generative Transformer architectures, such as T5 \citep{raffel2023exploringlimitstransferlearning} and BART \citep{lewis2019bartdenoisingsequencetosequencepretraining}, employing several decoding strategies for generating corrected versions of erroneous legal passages. 

In summary, our main contributions can be highlighted as follows:
\begin{itemize}
    \item We introduce a Romanian grammatical error taxonomy composed of 20 error types that best reflect specific Romanian error correction efforts, compiled from a set of common, widespread examples of Romanian grammatical errors extracted from several works of Romanian literature experts.
    \item We develop a synthetic error generation methodology comprising multiple text corruption methods, each designed to generate specific error types within our taxonomy. Our methods rely on a combination of statistical probabilities and Romanian language rules or specific characteristics to corrupt Romanian sentences and generate realistic and erroneous sequences.
    
    \item We present RoLegalGEC \footnote{https://huggingface.co/datasets/MirceaT/RoLegalGEC}, the first Romanian synthetic parallel dataset for error detection and correction in the legal domain, and evaluate the impact of our dataset by training a variety of GEC, as well as GED, architectures and analyzing the effects it has upon model performance, depending on factors such as architecture type, size, pre-training language, and output decoding strategy.

\end{itemize}

\section{Related work}

\paragraph{Legal NLP} NLP research in the legal domain so far has aimed to assist legal professionals by employing established neural architectures and training them on substantial data to interpret legal text and nuance accurately. Notably, LEGAL-BERT \citep{chalkidis-etal-2020-legal} is a collection of 
Bidirectional Encoder Representations from Transformers (BERT) \citep{devlin2019bert}, varying in size and training procedure, pre-trained on around 450,000 legal cases from Europe and the U.S., and applied to text classification and sequence tagging tasks. Furthermore, Lawma \citep{dominguezolmedo2025lawmapowerspecializationlegal} introduced 260 legal classification tasks tailored to specific legislative needs and applied LLaMa-based LLM approaches to annotated U.S. federal court databases, providing meaningful baselines for these tasks that, in addition to serving as standalone tools for law professionals, could be important for the further development of legal NLP tools. For Romanian, \cite{pais-etal-2021-named} used the comprehensive MARCELL-RO legal dataset \citep{tufics2020collection} and trained a bidirectional Long-Short Term Memory architecture \citep{hochreiter1997long} to detect specific entities in Romanian legislative texts, thus providing a baseline for the NER task and further enriching the NLP understanding of Romanian legal documentation.

\paragraph{Grammatical Error Taxonomies} Throughout research into GEC, the ERRANT \citep{bryant-etal-2017-automatic} has proven itself as the gold standard for error annotation, presenting a taxonomy consisting of 25 error types, ranging from general errors (e.g., spelling, punctuation, word order, etc.) to part-of-speech specific ones (e.g., noun errors, adjective form errors, subject-verb agreement errors, etc.). In terms of error detection, it addresses most requirements of the English language. Still, it presents a caveat by placing significant emphasis on parts of speech, only distinguishing a few different error types for each part of speech, a pitfall that research such as UA-GEC \citep{syvokon-etal-2023-ua} has aimed to remedy by proposing a Ukrainian taxonomy split into four categories (fluency, grammar, spelling, and punctuation) where, as an example, the "number" error aggregates incorrect uses of case for all notional parts of speech instead of splitting them between each specific part of speech. 

\paragraph{Synthetic Data Generation}

The process of synthetically generating realistic grammatical errors that mimic typical speaking or writing mistakes in a given language has been extensively studied for both high- and low-resource languages. For the English language, \citet{xu-etal-2019-erroneous} proposed an error generation procedure based on word trees that provide alternate context-erroneous versions of a word based on its root, as well as a rule-based, probability distribution-driven method that decides which tokens should be corrupted, as well as the nature of corruption, from a clean corpus sentence. This procedure is also described by the DirectNoise algorithm \citep{kiyono2019empiricalstudyincorporatingpseudo}, which injects erroneous noise into clean text based on statistics on how often token additions, deletions, and modifications occur in specialized settings (e.g., Wikipedia editing).
In low-resource settings, \citet{bondarenko-etal-2023-comparative} applied a taxonomy-based approach, aiming to replicate most of the errors in the UA-GEC taxonomy, by developing methodologies for generating punctuation, grammar, and fluency errors. Additionally, they studied the possibility of generating erroneous sentences via round-trip translation: translating the Ukrainian text into another language, namely Russian, and vice versa, thereby creating a likelihood of a different, and probably erroneous, version of the corpus example.  

An alternative approach to generating synthetic data, motivated by recent significant optimizations of LLMs and their enhanced ability to mimic reasoning skills, involves prompting LLMs to produce natural-sounding erroneous text when given a grammatically correct example as input.
\citet{stahlberg-kumar-2024-synthetic} constructed a tagged corruption model in which clean corpus sentences are assigned random error tags and then passed through a PaLM 2 LLM \citep{anil2023palm}, which reliably produces corrupted sentences according to the specified instructions. In addition, \citet{park2024chatlang8llmbasedsyntheticdata} generated sentence pairs from scratch, calling on the ChatGPT architecture \footnote{https://chat.openai.com/chat} to suggest possible subjects of a sentence along with possible specific grammatical errors, used to generate various sentences, both correct and erroneous, and ultimately constructing a chain-of-thought prompt that can match the generated sentences into parallel GEC sentence pairs.

\paragraph{Low-Resource Language GEC} \citet{bondarenko-etal-2023-comparative}, in addition to researching novel synthetic error generation procedures and developing comprehensive low-resource synthetic datasets, have also proposed several models suitable for solving the GEC task, either treating it as a machine translation task and employing a multilingual BART variant \citep{liu2020multilingual}, or using a pre-trained XML-RoBERTa model \citep{conneau2020unsupervised}, combined with a system of manual token-level transformations that aids in the matching of tokens between the parallel sentences in the GEC dataset, following the well-established GECToR methodology \citep{omelianchuk2020gectorgrammaticalerror}.

\citet{sharma-bhattacharyya-2025-indigec} proposed a novel solution for the masked language modeling aspect of a text-to-text Transformer, by translating the text, filling the gap, and then performing a back-translation to clear the mask. This technique is then applied alongside Transformer architectures, training on a manually annotated dataset of Indic-language examples to achieve an efficient low-resource language GEC tool. The authors also note improvements in model performance when using multilingual architectures over single-language ones.

For Turkish, \citet{kara2023gecturkgrammaticalerrorcorrection} developed a comprehensive rule-based synthetic generation procedure based on an official set of Turkish grammatical error rules, commissioned by the country's language association. A rule consists of prerequisite conditions that define the error, along with indications of how it can be corrected. The rules are thus treated as functions in a mathematical sense and have been implemented in code by the authors, who note that, to generate errors, they would have to implement the inverse functions on the clean corpus. Using the resulting datasets, a set of architectures has been implemented and evaluated for both error detection and correction tasks, ranging from a vanilla neural machine translation architecture to a BERT-based sequence tagging architecture and prefix tuning \citep{li2021prefix} on the mGPT \citep{shliazhko2024mgpt} infrastructure.

\paragraph{Romanian GEC} The established reference for grammatical error correction in Romanian, the RoGEC \citep{cotet} corpus employs vanilla Transformer architectures, initially trained on an artificial dataset, achieved by synthetically corrupting sentences extracted from Romanian Wikipedia articles in a probabilistic manner. The authors introduced a manually annotated parallel corpus of Romanian GEC gold-standard data, derived from an analysis of common spoken and written errors in Romanian mass media. They then use the corpus to further train the resulting Transformer model, either by resetting the initial training parameters or by fine-tuning while preserving parameters during gold-standard corpus training. More recently, \citet{info16030242} took into account both the detection and correction tasks by calling on a BERT encoder to detect erroneous subsequences in a given sentence, while a sequence-to-sequence model receives the erroneous sentence, along with markers consisting of special tokens that signal the error locations, ultimately outputting a corrected sentence. 

\section{Preliminaries}

\subsection{Task Formulation}

We focus our efforts towards three tasks: grammatical error detection (GED), the standard grammatical error correction task (GEC), and a correction task that also considers error tags resulting from GED as input for the correction model, which we will refer to as grammatical error correction based on detection (GEC-D).

\paragraph{Grammatical Error Detection} In our work, we treat GED as a token classification task. Thus, the input of the model will be a sentence \( x = (x_1,....x_n)\) that is suspected to contain grammatical errors, and the model will output a sequence of tags \( y = (y_1,....y_n)\), where a tag \(y_i\) is either a value out of the 20 error tags in our error taxonomy, signaling that the token situated in that specific location is erroneous and specifying the exact type of error it constitutes, or the 'O' tag, suggesting the fact that the token is grammatically correct.

\paragraph{Grammatical Error Correction}
We treat GEC as a sequence generation task using sequence-to-sequence architectures, since a GEC model takes a grammatically incorrect sentence as input and generates a grammatically correct sentence as output. Formally, the input of the GEC architecture is a structure \(U = \{X_1,...., X_n\}\). The target of the architecture is a structure \(V = \{Y_1,...., Y_n\}\), where \(X_i = \{x_{i1},....,x_{im}\}\) is a sequence of at least one word from the input sentence, and  \(Y_i = \{y_{i1},....,y_{ip}\}\) is a sequence of words that represents the grammatical correction of the sequence \(X_i\) (if the sequence \(X_i\) is already grammatically correct, we can assume \(X_i \equiv Y_i\), which means that the model will generate the same structure).

\paragraph{Grammatical Error Correction based on Detection}
We define the additional GEC-D task similarly to the GEC, as a sequence generation task, but instead of solely inputting the erroneous text, the input consists of both erroneous text and error tags. Formally, the model input will respect the structure \(U = \{X_1, X_2,...., X_n, <SEP>, t_1, t_2,...., t_n\}\), where \(X_i = \{x_{i1},....,x_{ip}\}\) represents a word token sequence in the erroneous sentence, and \(t_i = \{t_{i1},....,t_{ip}\}\) is the sequence of error tags associated with the token sequence  \(X_i\). Note that we use a special separator token to separate the word tokens from the tags. The output will respect the structure \(V = \{Y_1,....,Y_n\}\), where \(Y_i = \{y_{i1},....,y_{ip}\}\) is the correction of the sequence \(X_i\), guided by the error tag sequence \(t_i\).

\subsection{Synthetic Error Generation Methodology Preview}
The proposed Romanian error taxonomy defines a diverse set of error types; thus, we developed several synthetic error generation procedures to address the needs of each type, based on its nature, occurrence frequency, and reproduction complexity. In order to generate all the different types of error required for the detection and correction tasks, we managed to narrow down the methodology to three methods of synthetic error generation, which will be thoroughly detailed in the following sections, the methods being: 
\begin{itemize}
    \item Noise injection;
    \item Statistical generation based on confusion lists;
    \item LLM-based generation, further being split into:
    \begin{itemize}
        \item Zero-shot error generation;
        \item Two-shot error generation using common widespread Romanian error examples.
    \end{itemize}
\end{itemize}

Regarding LLM-based generation, the intention is to synthetically generate specific types of errors by prompting an LLM to analyze a sentence grammatically, thus minimizing the issue of non-annotated data, ultimately tasking it to generate an erroneous sentence that contains a specific error of choice, based on a clean corpus passage. For the purpose of choosing the most appropriate LLMs for this procedure, we experiment with general LLMs (i.e., GPT-4o, Meta LLaMa-3) and Romanian LLMs (i.e., RoMistral-7b, RoLLaMa2-7b) \citep{masala2024vorbeti} using Romanian or English language prompting, an example for each of the two prompting languages used on the same clean corpus sentence for the task of corrupting the sentence using the same type of error available in Figure~\ref{fig:promptlanguage}. Specifically, we test three setups: a Romanian-language prompt on the RoMistral-7b LLM, a Romanian-language prompt on the GPT-4o LLM, and an English-language prompt on the GPT-4o LLM. 

\begin{figure}[ht]
\centering
\footnotesize
\begin{minipage}{0.45\linewidth} 
\centering
\tiny
Vei juca rolul unui asistent în materie de gramatică a limbii române, care are obligația de a introduce erori gramaticale des întâlnite într-o propoziție corect gramaticală în limba română. Pentru sarcina de față, \textcolor{green}{va trebui să modifici în mod eronat gradul unui adjectiv} din următoarea propoziție : \textcolor{red}{<<Aprobată prin ORDINUL nr. 304 din 19 octombrie 2020, publicat în Monitorul Oficial al României, Partea I, nr. 1026 din 4 noiembrie 2020.>>}. Gândește-te cu mare atenție la rezolvarea acestei sarcini și analizează fiecare parte de vorbire în detaliu. Dacă respectiva parte de vorbire nu se regăsește în propoziție, răspunde strict cu NU. În caz contrar, răspunde strict cu noua propoziție eronată, urmată de INDICE: și indicii  pozițiilor în care ai modificat cuvinte față de propoziția inițială. \\[4pt]
\small
[Romanian language prompt]
\end{minipage}
\hfill
\begin{minipage}{0.45\linewidth} 
\centering
\tiny
You are a grammar assistant tasked with introducing common unacceptable errors within a correct Romanian language sentence. For this task, \textcolor{green}{you will have to change the degree of an adjective erroneously} in the following sentence: \textcolor{red}{<<Aprobată prin ORDINUL nr. 304 din 19 octombrie 2020, publicat în Monitorul Oficial al României, Partea I, nr. 1026 din 4 noiembrie 2020.>>}. Think very carefully about the task at hand, and analyze each part of speech in detail. If the specified part of speech does not exist in the sentence, strictly reply with NO. Otherwise, reply strictly with the erroneous sentence, followed by Index: and the positions of the words you have changed.\\[4pt]
\small
[English language prompt]
\end{minipage}
\caption{Side-by-side example of a Romanian language LLM prompt and an English language LLM prompt. The green text is formulated to indicate a specific error (in this case, the ADJ:FORM error). The red text represents the correct sentence from the clean corpus.}
\label{fig:promptlanguage}
\end{figure}

We tasked each selected combination of model and prompt language to generate errors and corrupt 10 clean corpus sentences, the outputs for each setup then being thoroughly manually analyzed in order to verify whether the LLM outputs align with the specific task given in the prompt, but also to observe how realistic and natural the generated errors sound for a native Romanian speaker. A showcase of the most representative example for our findings in manual analysis, which uses the prompts displayed in Figure~\ref{fig:promptlanguage} to attempt text corruption by changing the degree of an adjective, along with the LLM outputs for each of the experimental setups, is shown in Table~\ref{tab:llmlanguage}. 

\begin{table}[tbph]
    \centering
    \scriptsize
    \begin{tabular}{p{0.3\textwidth}p{0.6\textwidth}}
        \toprule
        \textbf{Prompting Setup} & \textbf{LLM Output} \\
        \midrule
        \midrule
         Romanian prompting on RoMistral-7b & "\textcolor{red}{Cea mai Aprobată} prin ORDINUL nr. 304 din 19 octombrie 2020, publicat în Monitorul Oficial al României, Partea I, nr. 1026 din 4 noiembrie 2020." 
        
        Indice: [0, 1, 2] \\
        \midrule
        Romanian prompting on GPT-4o & "Aprobată prin ORDINUL nr. 304 din 19 octombrie 2020, publicat în Monitorul \textcolor{red}{most Official} al României, Partea I, nr. 1026 din 4 noiembrie 2020." 
        
        Indice: [14, 15] \\
        \midrule
        English prompting on GPT-4o & "Aprobată prin ORDINUL nr. 304 din 19 octombrie 2020, publicat în Monitorul \textcolor{red}{cel mai Oficial} al României, Partea I, nr. 1026 din 4 noiembrie 2020." 
        
        Index: [14, 15, 16] \\
        \bottomrule
    \end{tabular}
    \caption{LLM output comparison between different LLM choices and prompt language setups, for the following clean corpus sentence: \textit{"Aprobată prin ORDINUL nr. 304 din 19 octombrie 2020, publicat în Monitorul Oficial al României, Partea I, nr. 1026 din 4 noiembrie 2020."}. The red text indicates the erroneous modification done by the LLM in order to corrupt the clean sentence.}
    \label{tab:llmlanguage}
\end{table}

The manual analysis we conducted highlights the fact that the Romanian RoMistral-7b LLM experiences difficulties in the process of identifying and distinguishing parts of speech in a sentence, which has resulted in most of the experiment samples not being corrupted by the LLM, even though the expected part of speech was present in the sentence, or offering misguided examples, such as the example in Table~\ref{tab:llmlanguage}. In this case, the word "Aprobată" (eng. "approved") is a participle that, while it may have an adjective form, it possesses the value of a verb, and thus, adjective degrees would not be allowed. Consequently, it creates a sentence that, while definitely corrupted and erroneous, acts upon the wrong part of speech; therefore, it compromises the accuracy of the GED error annotation process.

The Romanian prompting on the general GPT-4o LLM has no difficulty identifying parts of speech and performs well across most experimental samples. However, for some samples, this setup exhibits erratic behavior during the corruption phase, as shown in the example, where the Romanian prompt has led to the intrusion of English words into a Romanian sentence that, if translated into Romanian, would produce an accurate corruption.

Finally, the English prompting method on the GPT-4o LLM has performed best across all setups, showing no significant issues in the corruption and annotation processes and producing task-relevant erroneous sentences, as illustrated by the given example. Thus, to conclude, the findings of this experiment indicate that zero-shot and two-shot LLM prompting techniques will use the general-purpose, mainstream GPT-4o LLM in English.

\section{RoLegalGEC Dataset}

\subsection{Clean Corpora}
The most important prerequisite for creating a synthetic dataset for GED and GEC tasks is the grammatical correctness of the initial corpora used to generate corrupted examples. Thus, we propose that records of legal actions, as well as transcripts of high-level political proceedings, must be clear, coherent, and most importantly, grammatically correct so as not to leave any room for misinterpretation of the letter of the law. Hence, the two corpora we will process to create the RoLegalGEC dataset are MARCELL-RO \citep{varadi-etal-2020-marcell} and Europarl \citep{koehn2005europarl}.

The MARCELL-RO corpus is an aggregation of Romanian legislative documents emitted between 1990-2020, thus insuring grammatical contemporaneity, crawled from the public Romanian legislative web portal that have been grammatically analyzed by splitting and tokenizing each example, followed by a thorough annotation process that covers both the morphological and syntactical aspects of each token (e.g, lemmatization, part of speech, gender, number, and case). The Europarl corpus is a parallel corpus containing written records of European Parliament debates, translated into several languages, including Romanian. The choice to use both corpora aims to consider a variety of tones and modalities for expressing legal information, from the formal, written, and dense verbiage of the MARCELL-RO to the more informal yet still official discussion style of the Europarl. We analyze and process examples from these two corpora, resulting in a dataset of 350,000 examples. 

\subsection{Grammatical Error Taxonomy}

In order to facilitate RoLegalGEC annotation for the GED task and to properly evaluate models trained on this dataset, we introduce an error taxonomy tailored to Romanian. While conceptualizing the novel taxonomy, we have chosen to strike a balance between the rich complexity of the Romanian language and the intuitiveness and standardization of well-established error taxonomies, such as the ERRANT taxonomy \citep{bryant-etal-2017-automatic}. Thus, we propose a 20-error taxonomy that aims to cover most of the intricacies of Romanian,  while omitting errors we deem irrelevant from the English ERRANT taxonomy, taking into account specific aspects of Romanian grammar. All the different errors in our proposed taxonomy, together with a representative example for each error and their percentage share among all errors generated in the RoLegalGEC dataset, are listed in Table~\ref{tab:taxonomy}.    


\begin{table}[tbph]
\centering
\tiny
\begin{tabular}{lp{5cm}lp{1.9cm}cc}
\toprule
\textbf{Error Type} & \textbf{Brief Description} & \textbf{Romanian Language Example} & \textbf{Generation Method} & \textbf{Error Share} & \\
\midrule
ADJ & Inappropriate choice of adjective for the sentence context & [finalele] cinci partide $\rightarrow$ \textbf{[ultimele]} cinci partide & Two-shot prompting & 4.22\% \\
ADJ:FORM & Incorrect degree of an adjective & Muntele Everest este [înalt $\rightarrow$ \textbf{mai înalt}] ca Vf. Omu. & Zero-shot prompting & 2.94\% \\
ADV & Erroneous adverb usage & Am mâncat [decât $\rightarrow$ \textbf{doar}] migdale. & Two-shot prompting & 2.90\% \\
CONJ & Erroneous choice of conjunction & Am vrut [ca să $\rightarrow$ \textbf{să}] fim obiectivi. & Confusion list & 2.86\% \\
DET & Erroneous choice of determiner & El este [celui $\rightarrow$ \textbf{cel}] care primește mingea. & Confusion list & 1.69\% \\
MORPH & Misuse of words stemming from the same root & Consumați [minim $\rightarrow$ \textbf{minimum}] 2 litri de apă. & Two-shot prompting & 5.01\% \\
NOUN & Inappropriate noun usage & Le-am spus [la copii $\rightarrow$ \textbf{copiilor}] planul. & Two-shot prompting & 2.55\% \\
NOUN:INFL & Incorrect inflection form of plural noun & Mai ai și [succesuri $\rightarrow$ \textbf{succese}]. & Two-shot prompting & 1.34\% \\
NOUN:NUM & Incorrect number of a noun & Am comandat doi [hamburger $\rightarrow$ \textbf{hamburgeri}]. & Two-shot prompting & 1.58\% \\
NOUN:POSS & Disagreement between noun and possessive article & Poșeta este [a lui prietena $\rightarrow$ \textbf{a prietenei}] mele. & Zero-shot prompting & 0.39\% \\
ORTH & Incorrect use of whitespace that changes sentence meaning & Eu am [nu mai $\rightarrow$ \textbf{numai}] zece ani. & Noise injection & 10.52\% \\
PREP & Erroneous choice of preposition & Rămâneți [pe $\rightarrow$ \textbf{la}] telefon! & Confusion list & 3.13\% \\
PRON & Erroneous choice of pronoun & Tu [însuși $\rightarrow$ \textbf{însuți}] ai spus asta ieri. & Confusion list & 2.17\% \\
PUNCT & Inappropriate punctuation & Unde mergeți doamnă? $\rightarrow$ Unde mergeți[\textbf{,}] doamnă? & Confusion list & 10.76\% \\
SPELL & Errors related to word spelling & Am mers ieri la [dcotor $\rightarrow$ \textbf{doctor}]. & Noise injection & 25.55\% \\
VERB & Inappropriate choice of verb for the sentence context & Președinții se [urmează $\rightarrow$ \textbf{succedă}] o dată la 4 ani. & Two-shot prompting & 0.72\% \\
VERB:FORM & Erroneous choice of form in a verb & Nu [fă $\rightarrow$ \textbf{face}] și tu aceeași greșeală. & Two-shot prompting & 0.57\% \\
VERB:SVA & Disagreement between subject and verb in a sentence & Mi [s-a $\rightarrow$ \textbf{s-au}] făcut vrăji. & Two-shot prompting & 0.21\% \\
VERB:TENSE & Difference in tense between verb and rest of phrase & [Voi bănui $\rightarrow$ \textbf{bănuiesc}] că așa e politica. & Two-shot prompting & 0.38\% \\
WO & Incorrect word order & [Mai au $\rightarrow$ \textbf{Au mai}] rămas 15 minute. & Noise injection & 20.51\% \\
\bottomrule
\end{tabular}
\caption{The RoLegalGEC error taxonomy, along with examples adapted from \citet{texbook1} and \cite{texbook2}, the generation method used to inject the errors, and final error statistics.}
\label{tab:taxonomy}
\end{table}

\subsection{Common Romanian Language Grammatical Errors}
Creating a realistic synthetic GEC dataset that captures the intricacies of a rich language such as Romanian is virtually impossible without human-annotated data samples of error corrections by language experts. In general, for Romanian, there are very few such annotated datasets; one of the most notable is RoNACC \citep{cotet}, which focuses on collecting mass-media examples. Thus, we have undertaken research into Romanian grammar-oriented specialist literature, focusing on two books that highlight the most common grammatical mistakes made by native Romanian speakers \citep{texbook1, texbook2}. Based on the error examples provided in the books, along with the in-depth explanations provided by the authors regarding the nature of the erroneous utterances, we introduce a compact dataset of common Romanian grammatical errors which, as illustrated in Figure~\ref{fig:common}, consists of common erroneous utterances, along with expert corrections and the type of error as described in our own error taxonomy. Because this dataset does not include legal-domain examples, we will not use it directly to train our correction and detection models. Instead, it will serve as an essential tool for generating examples for the RoLegalGEC dataset, as will be illustrated below.

\begin{figure}[ht]
\centering
\footnotesize
\begin{minipage}{0.3\linewidth} 
\centering
Bună dimineața, \textcolor{red}{dragele} mele!\\[4pt]
[Erroneous sentence]
\end{minipage}
\hfill
\begin{minipage}{0.3\linewidth} 
\centering
Bună dimineața, \textcolor{green}{dragile} mele!\\[4pt]
\small
(eng. \textit{Good morning, my darlings!})\\[4pt]
[Corrected sentence]
\end{minipage}
\hfill
\begin{minipage}{0.3\linewidth} 
\centering
\footnotesize
O O O ADV O O \\[4pt]
[Error tag sequence]
\end{minipage}
\caption{Common Romanian language grammatical mistake example \citep{texbook1}, marked in red, along with the specialist-approved correction, marked in green, and potential error tag sequence.}
\label{fig:common}
\end{figure}

\subsection{Synthetic Error Generation Methods}

\paragraph{Noise Injection} This method represents the procedure intended to address general errors such as spelling, orthography, and word order errors, following an approach similar to the DirectNoise algorithm \citep{kiyono2019empiricalstudyincorporatingpseudo}. This procedure addresses two types of changes: word-level and character-level. For each of these levels, we consider four operations: substitution, deletion, insertion, and keeping. The detailed effects of each of the four operations for synthetic noise injection are given in Table~\ref{tab:rulebased}.

\begin{table}[thbp]
\centering
\footnotesize
\begin{tabular}{lp{7cm}p{6.7cm}}
\toprule
\textbf{Rule} & \textbf{Word-level Modification} & \textbf{Character-level Modification}\\
\midrule
Keep & Keep the current word unchanged &  Keep the current character unchanged \\[4pt]
Delete & Delete the current word & Delete the current character \\[4pt]
Insert & Insert a random word to the right of the current word OR bind the current word with the word to its right (ORTH) & Insert a random character to the right of the current character (ORTH) \\[12pt]
Substitute & Interchange the current word and the word to its right (WO) OR misspell the word (SPELL) & Interchange the current character and the character to its right OR change the current character to a random character OR change the current character in a controlled manner, by considering keyboard proximity, diacritics, or common Romanian misspellings. \\
\bottomrule
\end{tabular}
\caption{The effects of each noise injection rule for word and character level modifications. The resulting error tags are shown between parentheses.}
\label{tab:rulebased}
\end{table}

The procedure begins by identifying all non-punctuation and non-numeral tokens in a clean corpus sentence. For each such token, we probabilistically choose one of the four word-level operations. If the procedure decides that a token shall be misspelled, then each character of the token will be subject to one of the four character-level operations in the same probabilistic manner. To properly apply this procedure, we need to define the set of probabilities $ \{\mu_{substitution}, \mu_{deletion}, \mu_{insertion}, \mu_{keep}\}$.

\citet{kiyono2019empiricalstudyincorporatingpseudo}, along with subsequent papers that employ a similar statistical strategy \citep{tang2023hardsamplesrobusteffective}, deduce the set of probabilities by arbitrarily choosing a $\mu_{keep}$ value, then running an analysis of which $\mu_{substitution}$ value provides the best performance metric results, and ultimately assuming an equal number of deletions and insertions, meaning  $\mu_{deletion} = \mu_{insertion}$. 

We statistically support our chosen probability values by analyzing datasets of Wikipedia edit histories. The first notable dataset is the WikiAtomicEdits dataset \citep{DBLP:journals/corr/abs-1808-09422}, which analyzes token-level edits to Wikipedia articles across multiple languages and classifies them as additions or deletions. Unfortunately, there is no data on Romanian in this dataset; however, we can draw inspiration from statistics on several Romance languages (i.e., French, Spanish, and Italian). Comparing the number of token additions with the number of token deletions aggregated for these languages suggests a ratio of $\mu_{insertion}/\mu_{deletion} = 1.25$, rather than the ratio value of 1 suggested by the DirectNoise algorithm.

The optimal probability values discovered by \citet{kiyono2019empiricalstudyincorporatingpseudo} for substitutions (masks) and insertions are $\mu_{mask} = 0.5$ and $\mu_{insertion} = 0.15$, proposing a ratio of $\mu_{mask}/\mu_{insertion} = 3.33$. We will calibrate this ratio to our research by incorporating insights from another Wikipedia edit analysis source \citep{10.1371/journal.pone.0155305}, which focuses more on the editing behavior of Wikipedia contributors. Taking all factors into consideration, we deem a ratio of $\mu_{mask}/\mu_{insertion} = 3$ to be more appropriate for the injection of general, language-agnostic errors into the proposed clean legal sentences.

The final decision for the definition of our probability set concerns the value of $\mu_{keep}$. The DirectNoise algorithm authors consider an arbitrary value of $\mu_{keep} = 0.2$, a value intended to maximize erroneous text corruption, for the purpose of optimizing the efficiency with which GEC architectures identify and familiarize themselves with a wider range of errors. The downside of this choice is that it results in unrealistic, unnatural, and potentially unintelligible erroneous sequences, a caveat that we wish to avoid in the development of RoLegalGEC. 

Thus, in order to better simulate realistic errors for the purpose of generating natural erroneous sentences for our parallel dataset, we conduct a statistical analysis on the primary Romanian human-annotated parallel dataset for the GEC task, namely the RoNACC \citep{cotet}, which, as mentioned earlier, is a collection of manually gathered, analyzed, and corrected examples of errors occurring naturally in Romanian mass media. The statistical analysis consists of a word token assessment, with the aim of gauging the percentage of word tokens that remain unchanged between the parallel sentences relative to all available tokens. The results show that, throughout the whole RoNACC dataset, approximately 81.5\% of the words have not been altered between the erroneous and correct sentences in the sentence pairs, suggesting a $\mu_{keep}$ value of 0.815.

Considering various factors, such as the different niches of the datasets, the fact that the two aforementioned figures has been proposed and calculated for the whole dataset (whereas we only use this technique on a fraction of the errors in the taxonomy), or the need to adjust realistic dataset statistics to accommodate exploration of various error instances, we ultimately consider that a value of $\mu_{keep} = 0.7$ provides a good balance, allowing significant data corruption while maintaining a reasonable level of intelligibility and realism. Considering this value and the earlier defined ratios, we calculate the final probability values of the four defined operations for both word-level and character-level corruptions as being:

\begin{equation}
 \mu = \{\mu_{substitution}, \mu_{deletion}, \mu_{insertion}, \mu_{keep}\} = \{0.1875, 0.05, 0.0625, 0.7\} 
\end{equation}

\paragraph{Confusion List Generation} This procedure is applied to generate errors relating to the so-called "function" or "structure-class" words, namely CONJ, DET, PREP, and PRON errors; however, we will also be using it to generate PUNCT errors. In Romanian, there are a limited number of conjunctions, determiners, prepositions, and pronouns; therefore, for each of these parts of speech, we compiled a list of most, if not all, examples of that part of speech. The error generation procedure starts with the identification of all tokens that correspond to any of these four parts of speech for every clean corpus sentence. Then, we set an arbitrary probability of 30\%, which means that, for each identified token, there is a 30\% chance of substituting the token with a randomly chosen element from the confusion list corresponding to the speech part of the initial token. 

The compiled preposition list is as follows: "la"," în", "către", "contrar", "fără", "după", "cu", "lângă", "asupra", "de", "de la", "despre", "dimprejurul", "din", "dinaintea", "înspre", "între", "înăuntrul", "împotriva", "împrejurul", "înaintea", "înapoia", "întru", "dedesubtul", "datorită", "printre", "prin", "primprejur", "peste", "pentru", "pe", "până", "via", "spre", "sub". Suppose that the preposition "lângă" is encountered in a sentence, there is a 30\% likelihood that it will be replaced by another element of the preposition list, for example, "primprejur", and properly tagged as a PREP error. This procedure aims to create examples that are either grammatically correct but semantically inappropriate or both grammatically and semantically incorrect. 

\begin{figure}[tbph]
    \centering
    \includegraphics[width=0.45\textwidth]{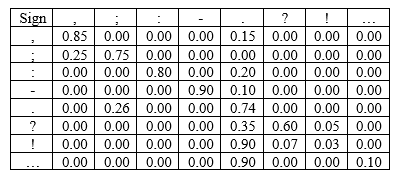}
    \caption{Punctuation transition probability matrix.}
    \label{fig:punct}
\end{figure}

However, the procedure for generating punctuation errors takes a different approach, inspired by a study of Ukrainian GEC \citep{bondarenko-etal-2023-comparative}, which requires defining a probabilistic transition matrix over the most common punctuation signs. We define our own transition matrix in Figure~\ref{fig:punct}. The generation procedure implies identifying all punctuation signs in a clean corpus sentence and passing them through the matrix that probabilistically decides which sign it outputs based on the input sign (e.g., a semicolon ";" has a 25\% chance of being turned into a colon "," and being tagged as a PUNCT error, and a 75\% chance of being kept the same).

\paragraph{Zero-Shot LLM Generation}
We generate the remaining errors in the taxonomy using LLM prompting techniques. The first procedure implies the application and the evaluation of the efficiency of zero-shot LLM prompting for certain types of errors that have proven to be tricky to generate using any other method, but also have a limited number of possible forms, such as possessive nouns and adjective form errors (where, for example, any adjective can only assume four degree forms). For these types of errors, we first choose which of the errors we want to inject into a sentence. Then, we feed a prompt to the GPT-4o LLM as exemplified in Table~\ref{tab:zeroshot}, where we assign the LLM to assume a role similar to an adversarial discriminator for Romanian grammar, having to analyze whether the input sentence possesses a part of speech that shall allow the corruption of the sentence using the error of choice, and ultimately require the LLM to generate the corrupted sentence following our guidelines, along with the index list of the tokens that have been modified. Processing the latter will allow us to deduce the error tag sequence of the corrupted sentence. 

\begin{table}[tbph]
    \centering
    \footnotesize
    \begin{tabular}{p{0.25\textwidth}p{0.7\textwidth}}
        \toprule
        \textbf{Prompting method component} & \textbf{Outcome} \\
        \midrule
        \midrule
        0-shot prompt & You are a grammar assistant tasked with introducing common unacceptable errors within a correct Romanian language sentence. For this task, \textcolor{green}{you will have to change the degree of an adjective erroneously} in the following sentence: \textcolor{red}{<<Aprobată prin ORDINUL nr. 304 din 19 octombrie 2020, publicat în Monitorul Oficial al României, Partea I, nr. 1026 din 4 noiembrie 2020.>>}. Think very carefully about the task at hand, and analyze each part of speech in detail. If the specified part of speech does not exist in the sentence, strictly reply with NO. Otherwise, reply strictly with the erroneous sentence, followed by Index: and the positions of the words you have changed." \\
        \midrule
        GPT-4o response & "Aprobată prin ORDINUL nr. 304 din 19 octombrie 2020, publicat în Monitorul cel mai Oficial al României, Partea I, nr. 1026 din 4 noiembrie 2020." 
        
        Index: [14, 15, 16] \\
        \midrule
        GED target & O O O O O O O O O O O O O O ADJ:FORM ADJ:FORM ADJ:FORM O O O O O O O O O O O O O O \\
        \bottomrule
    \end{tabular}
    \caption{Zero-shot prompt example. The green text is formulated to indicate a specific error (in this case, the ADJ:FORM error). The red text represents the correct sentence from the clean corpus.}
    \label{tab:zeroshot}
\end{table}

\paragraph{Two-Shot LLM Generation}
The two-shot LLM generation is the procedure intended for generating the rest of the challenging errors, ranging from specific errors of parts of speech (i.e., ADJ, ADV, NOUN, VERB), alternate forms of parts of speech (i.e., NOUN:INFL, NOUN:NUM, VERB:FORM, VERB:TENSE, VERB:SVA), to general morphology errors (i.e., MORPH). For each of these errors, we prompt the GPT-4o LLM using a few-shot technique, more precisely, a two-shot prompt. As illustrated in Table~\ref{tab:fewshot}, the prompt is similar to the one used for the zero-shot setting, except for the addition of two corruption examples of the error of choice, extracted from the common Romanian grammatical error dataset that we have previously defined. The LLM output has the same structure, generating the corrupted sentence and aiding the inference of its error-tag sequence. 

\begin{table}[!tbhp]
    \centering
    \scriptsize
    \begin{tabular}{p{0.2\textwidth}p{0.7\textwidth}}
\toprule
\textbf{Prompting method component} & \textbf{Outcome} \\
        \midrule
        \midrule
Few-shot prompt & You are a grammar assistant tasked with introducing common unacceptable errors in a correct Romanian sentence. For this task, \textcolor{green}{create a disagreement between a verb and a subject in a sentence, such that an erroneous sentence is created}.

For example, the sentence \textcolor{blue}{<<Mi s-au făcut vrăji.>>} will be turned into \textcolor{blue}{<<Mi s-a făcut vrăji.>>}

For example, the sentence \textcolor{blue}{<<Tinerii au luat cu asalt locația.>>} will be turned into \textcolor{blue}{<<Tinerii a luat cu asalt locația.>>}

So, the sentence \textcolor{red}{<<"La data de 19 februarie 2018, propunerea legislativă a fost prezentată în Biroul permanent al Camerei Deputaților și transmisă pentru raport și avize comisiilor de specialitate.">>} will be turned into \textcolor{red}{ANSWER}.

Think very carefully about the task at hand, and analyze each part of speech in detail. If the specified part of speech does not exist in the sentence, strictly reply with NO. Otherwise, reply strictly with the erroneous sentence, marked with ANSWER, followed by Index: and the positions of the words you have changed." \\
\midrule
GPT-4o response & "La data de 19 februarie 2018, propunerea legislativă a fost prezentată în Biroul permanent al Camerei Deputaților și transmisă pentru raport și avize comisiilor de specialitate."

Index: [11] \\
\midrule
GED target & O O O O O O O O O O O VERB:SVA O O O O O O O O O O O O O O O O \\
\bottomrule
\end{tabular}
\caption{2-shot prompt example. The green text represents the specific error (here, the VERB:SVA error), each error having a specific prompt. The red text is the corruptible sentence. The blue text shows the two-shot technique, which illustrates two common error examples of the specific error, adapted from \citet{texbook1} and drawn from the aforementioned example pool.}
\label{tab:fewshot}
\end{table}

We assign each of the ten errors addressed by this method a list of examples, consisting of erroneous sentences that exhibit the specific error, along with their correct counterparts; we refer to this set as a corruption example set (CES). The CES initially consists of instances of specific errors present in the common Romanian grammatical mistakes dataset. For the purpose of generating errors through the LLM two-shot prompting method, a process that is extensively detailed and visualized in Figure~\ref{fig:pipeline}, a two-step process is initiated:

\begin{itemize}
    \item The first step aims to enrich all corruption example sets. For each of the errors that would be dealt with by this few-shot technique, we construct several two-shot LLM prompt templates that contain two random examples from the specific error CES, and apply the prompts on several legal corpora sentences. We then manually analyze the LLM outputs and add the best resulting examples, consisting of erroneous-correct sentence pairs and the corresponding error tag sequence, to the specific error CES until the initial CES size is doubled. This step ensures that the final CES is not biased toward common errors and maintains a balance between common and legal errors.
    \item The second step involves processing the entire clean legal corpus, using the technique described above. For every sentence in the clean corpora, we assign an error type to inject into the sentence, then we choose two random samples from the enriched CES of the specific error, combining all these elements into two-shot prompts, designed to corrupt the whole dataset programmatically. The output is then processed similarly to the first step, transforming the index list into an error tag sequence that, together with the corrupted-correct sentence pair, defines a sample in the RoLegalGEC dataset. Using the two-shot prompting method in combination with, and in addition to, the other three generation procedures, we create the RoLegalGEC synthetic error generation dataset for error detection and correction.
\end{itemize}



\begin{figure}[ht]
\centering
\begin{tikzpicture}[node distance=1.6cm and 1.8cm]
\tikzstyle{every node}=[font=\scriptsize]
\node (ET) [cylinderstyle=red!20]{};
\node[below=0.1cm of ET] {Error taxonomy};
\node [boxstyle=red!20, right=1cm of ET] (ERR) {Error type};
\node [cylinderstyle=blue!20, above=3cm of ERR] (CRLM) {};
\node[above=0.1cm of CRLM, align=center] {Common error database};
\node [cylinderstyle=green!20, below=0.75cm of ERR] (CC) {};
\node[below=0.1cm of CC] {Clean corpus};
\node[boxstyle=green!20, right=1cm of CC] (CS) {Clean sentence};
\node[boxstyle=white!20, above=2.8cm of CS, text width = 1.2cm] (CES-X) {Extract CES for chosen error type};
\node[boxstyle=orange!20, right=1cm of CES-X, text width = 1cm, yshift=0.5cm] (CM1) {Common error 1};
\node[boxstyle=orange!20, below=0.5cm of CM1, text width = 1cm] (CM2) {Common error 2};
\node[boxstyle=white!20, right=4.7cm of ERR, text width = 2cm] (PROMPT) {Combining all elements into a prompt};
\node[boxstyle=purple!20, right=0.5cm of PROMPT] (LLM) {LLM};
\node[boxstyle=blue!20, right=0.75cm of LLM, text width = 1cm] (ES) {Erroneous sentence};
\node[boxstyle=white!20, right=0.05cm of LLM, yshift = 1cm, text width = 1cm] (IL) {Index list};
\node[boxstyle=yellow!20, right=0.5cm of IL, text width = 1cm] (ETS) {Error tag sequence};
\node[boxstyle=green!70, right=1.2cm of ES, text width=1cm] (PDE) {Parallel dataset example};
\node[diamondstyle, right=12cm of CRLM](CHECK){};
\node[above=0.1cm of CHECK, text width = 4cm, xshift=-1cm] {Check if manual analysis was passed and CES size < twice the initial size};
\node[boxstyle=white!20, right=6cm of CRLM, text width=2cm](CES-ADD) {Add generated example to CES};
\node[cylinderstyle=green!70, below=0.75cm of PDE](RLPD){};
\node[below=0.2cm of RLPD, text width=1.7cm] {RoLegalGEC};

\draw[arrowstyle] (ET) -- node[above, red, text width = 1cm, xshift=0.1cm]{Error choice} (ERR);
\draw[arrowstyle] (ERR) -- (CES-X);
\draw[arrowstyle] (CRLM) --(CES-X);
\draw[arrowstyle] (CES-X) -- node[above, red, text width = 2cm, yshift=0.45cm]{Random choice from error CES} (CM1);
\draw[arrowstyle] (CES-X) -- node[above, red, text width = 2cm, yshift=-0.95cm, xshift=-0.15cm]{Random choice from error CES} (CM2);
\draw[arrowstyle] (CC) -- (CS);
\draw[arrowstyle] (CS) -- (PROMPT);
\draw[arrowstyle] (CM1) -- (PROMPT);
\draw[arrowstyle] (CM2) -- (PROMPT);
\draw[arrowstyle] (ERR) -- (PROMPT);
\draw[arrowstyle] (PROMPT) -- node[above, red, text width = 1cm, xshift=0.25cm]{Input} (LLM);
\draw[arrowstyle] (LLM) -- node[above, red, text width = 1cm, xshift=0.1cm]{Output} (ES);
\draw[arrowstyle] (LLM) -- node[above, red, text width = 1cm, xshift=-0.2cm]{Output} (IL);
\draw[arrowstyle] (IL) -- node[above, red, text width = 1cm, yshift=0.25cm, xshift=-0.1cm]{Processing} (ETS);
\draw[arrowstyle] (ETS) -- (PDE);
\draw[arrowstyle] (CS) -- (PDE);
\draw[arrowstyle] (ES) -- (PDE);
\draw[arrowstyle] (PDE) -- node[above, red, text width = 2cm, xshift=0.2cm]{Manual analysis} (CHECK);
\draw[arrowstyle] (CHECK) -- node[above, red, text width = 2cm, xshift=0.75cm]{True} (CES-ADD);
\draw[arrowstyle] (CES-ADD) -- (CRLM);
\draw[arrowstyle] (PDE) -- node[above, red, text width = 2cm, yshift=-0.2cm, xshift=-0.25cm]{Add to final dataset}(RLPD);

\end{tikzpicture}
\caption{The LLM few-shot prompting error generation pipeline. When a clean sentence is extracted from the corpus and enters the pipeline, we choose an error to generate within based on the desired error shares (see Table~\ref{tab:taxonomy}). The common error database is then narrowed down to the CES of the specific, from which two error examples are extracted, and used along with the clean sentence and the formal formulation of the error type to construct an LLM prompt, as detailed in Table~\ref{tab:fewshot}. The LLM output consists of the corrupted sentence and the modification index list, the latter of which is processed into an error tag sequence. Both the clean and corrupted sentences, along with the tag sequence, make up a RoLegalGEC dataset example, which is added to the final dataset and analyzed to check if the necessary conditions are met so that the newly-generated example is included in its respective CES.}
\label{fig:pipeline}
\end{figure}

\subsection{Final Dataset}
The application of all four synthetic error generation methods on the two selected legal corpora has resulted in the first Romanian language legal synthetic dataset for the GED and GEC tasks, RoLegalGEC, a dataset consisting of 350,000 samples that, as illustrated in Figure~\ref{fig:datasetexample}, consist of:
\begin{itemize}
    \item A grammatically correct legal passage, extracted from the clean legal corpora;
    \item A grammatically incorrect passage, the result of the synthetic corruption of the correct legal passage;
    \item The error tag sequence, where each token of the erroneous passage is tagged based on whether it represents an error or not, along with the exact type of error.
\end{itemize}

        

\begin{figure}[ht]
\centering
\footnotesize
\begin{minipage}{0.47\linewidth} 
\raggedright
\textcolor{red}{Încât să}, măsurile asiguratorii au un caracter provizoriu, finalitatea lor constând în garantarea \textcolor{red}{exercităriiobligațiilor} cu caracter patrimonial, în cazul soluționării unui proces \textcolor{red}{peual}.\\[4pt]
{\centering [Erroneous sentence]\par}
\end{minipage}
\hfill
\begin{minipage}{0.47\linewidth} 
\raggedright
\textcolor{green}{Or}, măsurile asiguratorii au un caracter provizoriu, finalitatea lor constând în garantarea \textcolor{green}{exercitării obligațiilor} cu caracter patrimonial, în cazul soluționării unui proces \textcolor{green}{penal}.\\[4pt]
\small
(eng. \textit{However, precautionary measures are of a provisional nature, their purpose being to guarantee the exercise of patrimonial obligations, in the event of the resolution of a penal case.})\\
{\centering [Correct sentence]\par}
\end{minipage}
\par \vspace{16pt}
\begin{minipage}{0.9\linewidth} 
\centering
CONJ CONJ O O O O O O O O O O O O O ORTH O O O O O O O O O SPELL O \\[4pt]
[Error tag sequence]
\end{minipage}

\caption{RoLegalGEC dataset example. The erroneous sequences are highlighted in red, with their corrections highlighted in green.}
\label{fig:datasetexample}
\end{figure}

\subsection{Dataset Statistics}

The RoLegalGEC will be used to train, fine-tune, and evaluate a variety of error correction and detection models. The train-test split of this dataset will reflect the following guidelines:

\begin{itemize}
    \item The training dataset split consists of 315,000 examples, representing 90\% of the total, extracted, and corrupted from both the MARCELL-RO and Europarl corpora. Since the primary focus of the subsequent experiments is evaluation in the legal domain, we emphasize the predominance of examples from the MARCELL-RO corpus over those from Europarl. The contribution of each of the two clean corpora to the final training dataset is illustrated in Table~\ref{tab:datasetmetrics}. 
    \item The test dataset split will consist of the remaining 35,000 examples, 10\% of the total data, strictly extracted and processed from the MARCELL-RO legislative corpus, as we intend to evaluate solely the legal domain. The statistics for both the training and testing datasets are available in Table~\ref{tab:datasetmetrics}.
\end{itemize}

\begin{table}[tbph]
\centering
\small
\begin{tabularx}{0.88\linewidth}{lCCCCC}
\toprule 
Source Corpus & \# Sentences & \% of Dataset & \# Tokens & \# Err. Tokens & Error Rate \\
\midrule & \multicolumn{5}{c}{\textbf{Training Dataset}} \\
\midrule
MARCELL-RO & 197,214 & 62.60\% & 5,694,003 & 1,269,616 & 22.29\% \\  
Europarl & 117,786 & 37.40\% & 2,735,206 & 764,283 & 27.94\% \\
Total & 315,000 & 100.00\% & 8,429,209 & 2,033,899 & 24.13\% \\
\midrule & \multicolumn{5}{c}{\textbf{Test Dataset}} \\
\midrule
MARCELL-RO & 35,000 & 100.00\% & 1,007,420 & 214,470 & 21.28\% \\
\bottomrule
\end{tabularx}
\caption{RoLegalGEC dataset composition statistics.}
\label{tab:datasetmetrics}
\end{table}

\section{Method}

 \subsection{Grammatical Error Detection Architectures}

\paragraph{DistilBERT}
Given that we are dealing with token classification, the first technique we intend to employ for the GED task is to fine-tune a general-purpose pre-trained encoder architecture for the downstream task of error annotation on the RoLegalGEC dataset. Also, given the narrow nature of the task at hand, as implied by its definition and the limited set of error tags, we make the practical choice to fine-tune and evaluate DistilBERT architectures \citep{sanh2020distilbertdistilledversionbert} for the GED task, as they are lighter and more efficient encoders obtained through the knowledge distillation \citep{hinton2015distilling} of classic BERT encoders. We will be relying on the original multilingual DistilBERT (mDistilBERT) model \citep{sanh2020distilbertdistilledversionbert}, pre-trained in over 100 languages, including Romanian, as well as experimenting with a Romanian DistilBERT (RoDistilBERT) model \footnote{https://huggingface.co/racai/distilbert-base-romanian-cased} \citep{avram2022distillingknowledgeromanianberts}, to determine whether it can provide more reliable error tag deductions than the original model.

\paragraph{Sequence Tagging}
Following the impressive results of replicating the GECToR formula for low-resource languages, such as  Turkish \citep{kara2023gecturkgrammaticalerrorcorrection}, we also consider tackling the GED task by using a sequence tagging architecture that will be adapted to Romanian. This method is going to take a slightly different approach to the DistilBERT methods because, while the DistilBERT decoder would solely treat this task as a token classification task, the sequence tagger acts more like a machine translation task, since it uses a BERT model as an encoder that can predict if a token needs to be changed, along with the nature of the modification, should that be necessary, and thus predict the tag of the token accordingly. In contrast, the decoder is a simple softmax layer that outputs only the appropriate tags, not the entire corrected sentence, to achieve the ultimate goal of the defined GED task.  

For the purpose of error detection and annotation, we will experiment with two variants of a Romanian sequence tagger, both using the Romanian BERT (RoBERT) architecture \citep{masala-etal-2020-robert}, pre-trained on 12.6 GB of Romanian corpora, as the baseline encoder, complemented by softmax-layer decoders. The two Sequence Tagging variants are SeqTag-base, which employs a base version of the Romanian BERT (RoBERT-base) as encoder, and SeqTag-large, which uses a larger and more complex variant of the Romanian BERT (RoBERT-large) as encoder.

\subsection{Grammatical Error Correction Architectures}

\paragraph{BART} Given its proven proficiency in text generation tasks across multiple metrics and benchmarks, the first experiments on error correction tasks involve fine-tuning three BART architectures \citep{lewis2019bartdenoisingsequencetosequencepretraining}, namely two size variants of the BART architecture that have been pre-trained for Romanian: a base variant (RoBART-base) \footnote{https://huggingface.co/Iulian277/ro-bart-512} and a large variant (RoBART-large), along with one multilingual BART variant (mBART) \citep{liu2020multilingual}. For the GEC task, we extract \(\{X, Y\}\) tuples from the training dataset, \(X\) being the input erroneous sentence and \(Y\) being the output corrected sentence to feed into a classic BART architecture. 

For the GEC-D task, we create tag-augmented BART models, starting from each of the Romanian pre-trained or multilingual BART architectures, by modifying the input structure and, implicitly, the models' objective function to enable the inclusion of both the erroneous sentence and the error tag sequence as model inputs. For this task, the whole extent \(\{X, T, Y\}\) of the RoLegalGEC dataset will be used, as the erroneous sequence \(X\) and the error tag sequence \(T\) will be processed, pasted and delimited to create an intelligible input, that will be fed to the BART models along with the output correct sentence \(Y\). 

\paragraph{T5} The second architecture that will constitute the basis for researching and evaluating the Romanian language error correction for the legal domain is an architecture praised for its versatility in the field of text generation, namely the T5 architecture \citep{raffel2023exploringlimitstransferlearning}, which uses concepts of transfer learning to generate reliable text outputs for multiple tasks in multiple domains. Similar to the BART experiments, the GEC mechanism will start with three pre-trained T5 model variants, two Romanian models: a base variant (RoT5-base) \footnote{https://huggingface.co/dumitrescustefan/t5-v1\_1-base-romanian} and a large variant (RoT5-large), along with one multilingual variant (mT5) \citep{xue-etal-2021-mt5}, which will all be further fine-tuned for the GEC task using erroneous-correct sentence pairs in the training dataset.

Similar to BART, we develop tag-augmented RoT5-base, RoT5-large, and mT5 architectures for training and evaluating the GEC-D task, applying modifications that enable the use of erroneous text with its corresponding error annotation as input for text correction. 

\section{Evaluation}

\subsection{Performance Metrics}

\paragraph{Grammatical Error Correction} Given the fact that we are evaluating sequence-to-sequence generative architectures, the GEC evaluation, as well as the GEC-D, has to account for the manner in which the chosen decoding strategy of the model decoder impacts the overall performance of the model. Thus, for each architecture, we consider the evaluation of two decoding strategies, namely:

\begin{itemize}
    \item Top-p search - considers the tokens with the highest likelihoods, whose aggregated probabilities do not exceed a p-value (in our case, p-value = 0.9), ultimately generating one of these tokens at random.
    \item Beam search - considers the most likely B tokens (in our case, B = 5). The search space for the following token consists of the most likely B tokens to be generated after each of the initial B tokens, and so on.
\end{itemize}

Ultimately, the correction models for both the GEC and GEC-D tasks will be evaluated based on the precision, recall, and \(F_{0.5}\) score provided by the MaxMatch scorer \citep{dahlmeier-ng-2012-better}.

\paragraph{Grammatical Error Detection} Following our own error taxonomy, we evaluate the GED task and keep a confusion matrix for each error type. After model inference, each dataset example will benefit from a predicted error tag sequence and its corresponding target error tag sequence. Both sequences will be iterated through and compared tag by tag. If we predict an error tag, the target tag can either match the predicted tag, in which case it is tallied as a True Positive for the predicted tag, or differ from it, in which case it is tallied as a False Positive for the predicted tag. Alternatively, if we predict a correct token when the target is an error tag, we count it as a False Negative for the target error tag.

Given the outlined confusion matrices, the GED models will be evaluated based on the precision, recall, and \(F_{0.5}\) scores for each of the 20 errors in the taxonomy, as well as the overall precision, recall, and \(F_{0.5}\) scores resulting from the aggregated confusion matrix of all individual errors. 

\subsection{Experimental Details}

All of our proposed experiments for the GED, GEC, or GEC-D tasks use a baseline specialized pre-trained architecture, as described in the Final Dataset section, which is then further trained on the training split of the RoLegalGEC dataset. All experiments employ a cross-entropy loss function to monitor the training procedure. 

\paragraph{Grammatical Error Detection Experiments}
As for the DistilBERT GED experiments, the multilingual training is based on the pre-trained mDistilBERT architecture, a 134M-parameter model, which we further train on the RoLegalGEC training dataset for a duration of 3 epochs, employing a learning rate of 3e-5. In contrast, the RoDistilBERT architecture, a more compact model with approximately 82M parameters, required more training for 4 epochs with a learning rate of 5e-5; both training setups performed optimally with a weight decay value of 0.01.  

Regarding the choice of encoder for the sequence tagging GED architecture experiments, we focus exclusively on Romanian pre-trained encoders, experimenting with two different-sized variants of the RoBERT architecture. The SeqTag-base model uses the RoBERT-base encoder, which has 12 encoder layers, with a hidden size of 768 and 12 attention heads, totaling 114M parameters. At the same time, the SeqTag-large model uses the more complex RoBERT-large encoder, with 24 encoder layers, a hidden size of 1024, 24 attention heads, and 341M parameters.

Both the SeqTag-base and SeqTag-large sequence architectures, trained in the same experimental setup, appear to perform optimally when trained on the RoLegalGEC training dataset for 3 epochs, a learning rate of 2e-5, and a weight decay of 0.01.

\paragraph{Grammatical Error Correction Experiments}

As previously mentioned, for the purpose of experimenting with the BART architecture for correction tasks, we experiment with three baseline BART models: a standard multilingual BART architecture (mBART) and a base size Romanian pre-trained BART model (RoBART-base), both with 140M training parameters, along with a large variant of the Romanian BART model (RoBART-large) \footnote{https://huggingface.co/Iulian277/ro-bart-large-512}, having 400M parameters. For the GEC sub-task, all three models, the mBART, RoBART-base, and RoBART-large, require 5 training epochs, whereas for the GEC-D task, 4 epochs are sufficient for all three architectures.

We set the learning rate to 3e-5 for the GEC mBART, GEC RoBART-base, and GEC-D RoBART-base setups, while the GEC-D mBART, GEC RoBART-large, and GEC-D RoBART-large setups use a learning rate of 2e-5; none of the experiments involving the BART architecture for error correction employ any weight decay.

Similar to the BART experiments, the T5 experimental setup involves the mT5, RoT5-base, and RoT5-large \footnote{https://huggingface.co/dumitrescustefan/t5-v1\_1-large-romanian} architectures, having 223M, 247M, and 783M parameters, respectively. All experiments, across all three architectures and both GEC and GEC-D environments, required 3 training epochs with a weight decay of 0.02. However, the GEC RoT5-base and GEC-D RoT5-large models performed optimally at a learning rate of 2e-5, whereas the mT5 and RoT5-large models trained in the GEC environment performed best at a value of 3e-5. In addition, both the GEC-D mT5 and RoT5-base models performed best at a learning rate of 4e-5.

\section{Results}

\paragraph{Grammatical Error Detection} The previously detailed detection architectures, namely the DistilBERT models and the Sequence Tagger variants, along with their respective training setups, have been trained using the RoLegalGEC training dataset. The results of evaluating the models on the RoLegalGEC test dataset for each error type, as well as the overall performance metrics, are presented in Table~\ref{tab:gedmetrics}. 

\begin{table}[tbph]
\centering
\footnotesize
\begin{tabularx}{\linewidth}{lCCCCCCCCCCCC}
\toprule
 & \multicolumn{3}{c}{\textbf{RoDistilBERT}}  & \multicolumn{3}{c}{\textbf{mDistilBERT}} & \multicolumn{3}{c}{\textbf{SeqTag-base}} & \multicolumn{3}{c}{\textbf{SeqTag-large}}\\
\cmidrule(lr){2-4} \cmidrule(lr){5-7} \cmidrule(lr){8-10} \cmidrule(lr){11-13}
Error & P & R & $F_{0.5}$ & P & R & $F_{0.5}$ & P & R & $F_{0.5}$ & P & R & $F_{0.5}$\\
\cmidrule(lr){1-4} \cmidrule(lr){5-7} \cmidrule(lr){8-10} \cmidrule(lr){11-13}
ADJ & 0.681 & 0.638 & \textbf{0.672} & 0.350 & 0.261 & 0.327 & 0.473 & 0.256 & 0.404 & 0.532 & 0.610 & 0.546 \\
ADJ:FORM & 0.734 & 0.595 & 0.701 & 0.744 & 0.758 & \textbf{0.747} & 0.674 & 0.473 & 0.621 & 0.713 & 0.588 & 0.684 \\
ADV & 0.773 & 0.719 & 0.762 & 0.758 & 0.782 & 0.763 & 0.755 & 0.851 & 0.773 & 0.781 & 0.817 & \textbf{0.788} \\
CONJ & 0.945 & 0.752 & 0.899 & 0.880 & 0.767 & 0.855 & 0.923 & 0.799 & 0.895 & 0.914 & 0.852 & \textbf{0.901} \\
DET & 0.817 & 0.846 & 0.823 & 0.915 & 0.875 & \textbf{0.907} & 0.894 & 0.854 & 0.886 & 0.841 & 0.807 & 0.834 \\
MORPH & 0.816 & 0.831 & 0.819 & 0.846 & 0.800 & \textbf{0.837} & 0.725 & 0.732 & 0.727 & 0.738 & 0.686 & 0.727 \\
NOUN & 0.750 & 0.787 & \textbf{0.757} & 0.636 & 0.806 & 0.664 & 0.641 & 0.825 & 0.670 & 0.661 & 0.743 & 0.676 \\
NOUN:INFL & 0.525 & 0.573 & 0.534 & 0.577 & 0.718 & 0.601 & 0.708 & 0.379 & 0.603 & 0.701 & 0.572 & \textbf{0.671} \\
NOUN:NUM & 0.792 & 0.803 & 0.795 & 0.876 & 0.857 & 0.872 & 0.726 & 0.875 & 0.752 & 0.861 & 0.924 & \textbf{0.873} \\
NOUN:POSS & 0.841 & 0.670 & 0.800 & 0.767 & 0.643 & 0.739 & 0.891 & 0.703 & 0.846 & 0.887 & 0.803 & \textbf{0.869} \\
ORTH & 0.833 & 0.869 & 0.840 & 0.881 & 0.910 & 0.886 & 0.920 & 0.816 & 0.897 & 0.911 & 0.849 & \textbf{0.898} \\
PREP & 0.463 & 0.757 & \textbf{0.502} & 0.515 & 0.446 & 0.499 & 0.119 & 0.218 & 0.131 & 0.524 & 0.398 & 0.493 \\
PRON & 0.820 & 0.382 & 0.667 & 0.868 & 0.744 & \textbf{0.840} & 0.642 & 0.126 & 0.354 & 0.722 & 0.670 & 0.711 \\
PUNCT & 0.766 & 0.554 & 0.712 & 0.789 & 0.863 & \textbf{0.803} & 0.689 & 0.915 & 0.724 & 0.734 & 0.677 & 0.722 \\
SPELL & 0.862 & 0.651 & 0.809 & 0.857 & 0.726 & 0.827 & 0.875 & 0.831 & 0.866 & 0.869 & 0.879 & \textbf{0.871} \\
VERB & 0.815 & 0.725 & 0.795 & 0.870 & 0.688 & 0.826 & 0.890 & 0.722 & \textbf{0.850} & 0.813 & 0.871 & 0.824 \\
VERB:FORM & 0.690 & 0.719 & 0.695 & 0.807 & 0.738 & \textbf{0.792} & 0.690 & 0.755 & 0.702 & 0.757 & 0.832 & 0.771 \\
VERB:SVA & 0.892 & 0.805 & 0.873 & 0.788 & 0.930 & 0.813 & 0.928 & 0.777 & 0.893 & 0.920 & 0.886 & \textbf{0.913} \\
VERB:TENSE & 0.701 & 0.658 & 0.692 & 0.717 & 0.701 & 0.714 & 0.696 & 0.728 & 0.702 & 0.714 & 0.719 & \textbf{0.715} \\
WO & 0.777 & 0.735 & 0.768 & 0.847 & 0.869 & \textbf{0.852} 
& 0.677 & 0.656 & 0.673 & 0.671 & 0.676 & 0.672 \\
\cmidrule(lr){1-4} \cmidrule(lr){5-7} \cmidrule(lr){8-10} \cmidrule(lr){11-13}
\textbf{Average} & \textbf{0.797} & \textbf{0.679} & \textbf{0.770} & \textbf{0.817} & \textbf{0.774} & \textbf{0.808} & \textbf{0.786} & \textbf{0.742} & \textbf{0.777} & \textbf{0.785} & \textbf{0.766} & \textbf{0.781} \\

\bottomrule
\end{tabularx}
\caption{Performance metrics for GED tasks on the RoLegalGEC dataset.}
\label{tab:gedmetrics}
\end{table}

The results show that the DistilBERT models, particularly RoDistilBERT, specialize in detecting more general errors related to parts of speech (especially ADJ and NOUN, but also ADV and VERB). In contrast, the Sequence Tagging models are better suited to annotate errors relating to particular characteristics of parts of speech (e.g., NOUN:INFL, NOUN:NUM, NOUN:POSS, VERB:SVA, VERB:TENSE). Additionally, the structure-class word errors we generated using the confusion list method (CONJ, DET, PREP, PRON) seem to be overall better identified by the DistilBERT models. The CONJ and DET errors are exceptionally well annotated by all models, whereas the PREP and PRON errors pose challenges for the SeqTag models, significantly reducing their performance relative to the DistilBERT models. 

Finally, the expectation is that a BERT-based Sequence Tagging architecture would be more proficient in language-agnostic error detection than a knowledge-distilled BERT architecture. While this hypothesis has a basis highlighted by the slight edge that the SeqTag-large model has over DistilBERT models for spelling and orthography errors, the complete results lead to the conclusion that the DistilBERT models, especially the mDistilBERT, obtain the best results on our RoLegalGEC dataset for these types of errors, exemplified by the proficiency in detecting morphology, punctuation and, most notably, word order errors. 

Concluding the analysis of the metric results in Table~\ref{tab:gedmetrics}, we deduce that, even though the SeqTag-large model has the highest number of taxonomy errors that it is proficient in, the mDistilBERT emerges as the best model overall, due to the higher frequency and relevance of the errors that the latter model specializes in. Overall, the mDistilBERT obtains an impressive 0.81 \(F_{0.5}\) score on the RoLegalGEC test dataset, slightly outperforming the State-of-the-Art for Romanian GED when it comes to the most relevant errors shared between the two error taxonomies involved; however, our optimal \(F_{0.5}\) score of 0.81 falls short of the state-of-the-art for low-resource languages as a whole by about 0.11, differences that can also be attributed to the distinct evaluation methodologies, as well as the difference in nature of the test datasets between all separate research experiments.

\paragraph{Grammatical Error Correction}
Both sequence-to-sequence generative architectures, BART and T5, along with their different size and pre-training language variants, have been fine-tuned using the RoLegalGEC training dataset, with the primary aim of assessing which architecture is better equipped to handle the erroneous nature of the synthetic legal dataset and to properly generate corrected sentences. Additionally, we will observe the impact of each specified decoding strategy on the performance of each model variant of our two main architectures.

The T5 and BART performance metrics for the Romanian base, large, and multilingual model variants, in both classic and tag-augmented setups, are available in Table~\ref{tab:gecmetrics}. Later in the Discussion section, the metric results will prove helpful, along with relevant inference examples, in providing an answer to our second research question, whether supplying error tags in the training process improves the correction capabilities of the generative architectures, as well as partially answering our third research question, by analyzing the manner in which the model size and the pre-training language choice impacts model performance for the correction tasks.  

\begin{table}[tbph]
\centering
\small
\begin{tabular}{lcccccc}
\toprule
 & \multicolumn{3}{c}{Top-p search} & \multicolumn{3}{c}{Beam search } \\
 & \multicolumn{6}{c}{\textbf{Grammatical Error Correction (GEC)}} \\
\cmidrule(lr){1-4} \cmidrule(lr){5-7}
Model
& P & R & $F_{0.5}$
& P & R & $F_{0.5}$ \\
\cmidrule(lr){1-4} \cmidrule(lr){5-7}
RoBART-base
& 0.478 & 0.432 & 0.468
& 0.505 & 0.464 & 0.496 \\
RoT5-base
& 0.514 & \textbf{0.449} & 0.499
& 0.563 & \textbf{0.506} & 0.551 \\
RoBART-large
& 0.478 & 0.386 & 0.456
& 0.503 & 0.432 & 0.487 \\
RoT5-large
& 0.550 & 0.441 & \textbf{0.524}
& 0.593 & 0.443 & \textbf{0.556} \\
mBART
& 0.521 & 0.435 & 0.501
& 0.541 & 0.456 & 0.521 \\
mT5
& \textbf{0.557} & 0.404 & 0.518
& \textbf{0.597} & 0.419 & 0.551 \\
\cmidrule(lr){1-4} \cmidrule(lr){5-7}
 & \multicolumn{6}{c}{\textbf{Tag-augmented Error Correction (GEC-D)}} \\
\cmidrule(lr){1-4} \cmidrule(lr){5-7}
Model
& P & R & $F_{0.5}$
& P & R & $F_{0.5}$ \\
\cmidrule(lr){1-4} \cmidrule(lr){5-7}
RoBART-base
& 0.425 & 0.485 & 0.436
& 0.528 & 0.481 & 0.518 \\
RoT5-base
& 0.519 & \textbf{0.509} & 0.517
& 0.569 & 0.518 & \textbf{0.558} \\
RoBART-large
& 0.446 & 0.382 & 0.432
& 0.535 & 0.447 & 0.515 \\
RoT5-large
& 0.537 & 0.503 & 0.530
& 0.562 & \textbf{0.532} & 0.556 \\
mBART
& 0.493 & 0.409 & 0.474
& 0.539 & 0.454 & 0.520 \\
mT5
& \textbf{0.585} & 0.408 & \textbf{0.538}
& \textbf{0.617} & 0.397 & 0.556 \\
\bottomrule
\end{tabular}
\caption{Performance metrics for GEC and GEC-D correction tasks on the RoLegalGEC.}
\label{tab:gecmetrics}
\end{table}

While the BART architecture is known for its proficiency in handling ambiguous masked language in a human-like manner, our experiments prove that, for the task of legal Romanian GEC on the novel RoLegalGEC dataset, the ability of the T5 architecture to adapt to the task at hand using prior knowledge renders T5-based model variants as the most efficient correction tools throughout all of the evaluation setups. As far as decoding strategies go, evaluation results show that the enhanced search space provided by the beam search vastly outperforms the simplicity of the top-p search. Based on these findings, corroborated by the metric results, the BART-top-p setting stands out as inefficient, whereas pairing the T5 architecture with the beam search strategy seems to be the most beneficial for the correction tasks. 

After analyzing the metric performance of all different size and language variants of T5 and BART architectures for correcting RoLegalGEC dataset sentences, we conclude that the RoT5-large model variant paired with the beam search decoding strategy provides the best results in a classic GEC setting. Still, the overall optimal setup appears to be the beam search RoT5-base model in a tag-augmented GEC-D setup. Both setups bring the metric results above the 0.55 \(F_{0.5}\) score mark and manage to push the performance of our models past the current State-of-the-art models for the Romanian GEC, proving to be far more efficient than experiments employing vanilla NMT architectures, and trailing 0.07 behind low-resource language Transformer-based model State-of-the-Art and 0.16 to methods involving GPT prefix tuning architectures, taking into consideration once again the different evaluation setups for the distinct experiments.

\section{Discussion}

\subsection{Response for RQ1}

To answer this question, we conduct an experiment where we intend to corrupt 12 clean sentences, one for each of the 12 error types in the taxonomy that are not language-agnostic, by applying a zero-shot LLM prompt. Meanwhile, we also attempt to corrupt them using a two-shot LLM prompt based on common Romanian mistakes representative of the error type. Finally, we conduct a manual analysis to verify the corruption accuracy of both prompting methods on the GPT-4o LLM. The expectation is a trade-off between the higher corruption accuracy achieved through the extended context information provided by the two-shot prompting method and the time and resource efficiency of the zero-shot prompting technique. 

\begin{table}[tbph]
    \centering
    \tiny
    \begin{tabular}{p{0.05\textwidth}p{0.65\textwidth}p{0.2\textwidth}}
        \toprule
        \textbf{Technique} & \textbf{Prompt used} & \textbf{GPT-4o LLM Output} \\
        \midrule
        \midrule
        Zero-shot & You are a grammar assistant tasked with introducing common unacceptable errors within a correct Romanian language sentence. For this task, \textcolor{green}{create a disagreement between a verb and a subject in a sentence, such that an erroneous sentence is created}, in the following sentence: \textcolor{red}{<<"La data de 19 februarie 2018, propunerea legislativă a fost prezentată în Biroul permanent al Camerei Deputaților și transmisă pentru raport și avize comisiilor de specialitate.">>}.

Think very carefully about the task at hand, and analyze each part of speech in detail. If the specified part of speech does not exist in the sentence, strictly reply with NO. Otherwise, reply strictly with the erroneous sentence, followed by Index: and the positions of the words you have changed. & "La data de 19 februarie 2018, propunerea \textbf{legislativ} a fost prezentată în Biroul permanent al Camerei Deputaților și transmisă pentru raport și avize comisiilor de specialitate."

Index: [7] \\
        \midrule
        Few-shot & You are a grammar assistant tasked with introducing common unacceptable errors within a correct Romanian language sentence. For this task, \textcolor{green}{create a disagreement between a verb and a subject in a sentence, such that an erroneous sentence is created}.

For example, the sentence \textcolor{blue}{<<Mi s-au făcut vrăji.>>} will be turned into \textcolor{blue}{<<Mi s-a făcut vrăji.>>}

For example, the sentence \textcolor{blue}{<<Tinerii au luat cu asalt locația.>>} will be turned into \textcolor{blue}{<<Tinerii a luat cu asalt locația.>>}

So, the sentence \textcolor{red}{<<"La data de 19 februarie 2018, propunerea legislativă a fost prezentată în Biroul permanent al Camerei Deputaților și transmisă pentru raport și avize comisiilor de specialitate.">>} will be turned into \textcolor{red}{ANSWER}.

Think very carefully about the task at hand, and analyze each part of speech in detail. If the specified part of speech does not exist in the sentence, strictly reply with NO. Otherwise, reply strictly with the erroneous sentence, marked with ANSWER, followed by Index: and the positions of the words you have changed. & "La data de 19 februarie 2018, propunerea legislativă a fost \textbf{prezentat} în Biroul permanent al Camerei Deputaților și transmisă pentru raport și avize comisiilor de specialitate."

Index: [11] \\
        \midrule
        \midrule
        Zero-shot & You are a grammar assistant tasked with introducing common unacceptable errors within a correct Romanian language sentence. For this task, \textcolor{green}{change the inflection of one of the nouns in the following sentence, in order for the sentence to be grammatically incorrect}: \textcolor{red}{<<Președinții celor două Camere ale Parlamentului, Guvernul și Avocatul Poporului nu au comunicat punctele lor de vedere asupra excepției de neconstituționalitate.>>}. Think very carefully about the task at hand, and analyze each part of speech in detail. If the specified part of speech does not exist in the sentence, strictly reply with NO. Otherwise, reply strictly with the erroneous sentence, followed by Index: and the positions of the words you have changed. & "Președinții celor două Camere ale Parlamentului, Guvernul și Avocatul Poporului nu au \textbf{comunicatul} punctele lor de vedere asupra excepției de neconstituționalitate."

        Index: [13] \\
        \midrule
        Few-shot & You are a grammar assistant tasked with introducing common unacceptable errors within a correct Romanian language sentence. For this task, \textcolor{green}{change the inflection of one of the nouns in a sentence, in order for the sentence to be grammatically incorrect}. For example, the sentence \textcolor{blue}{<<Mai ai și eșecuri, mai ai și succese.>>} will be turned into \textcolor{blue}{<<Mai ai și eșecuri, mai ai și succesuri.>>}. For example, the sentence \textcolor{blue}{<<Am văzut trei cuțite frumoase de vânzare.>>} will be turned into \textcolor{blue}{<<Am văzut trei cuțituri frumoase de vânzare.>>}. So, the sentence: \textcolor{red}{<<Președinții celor două Camere ale Parlamentului, Guvernul și Avocatul Poporului nu au comunicat punctele lor de vedere asupra excepției de neconstituționalitate.>>} will be turned into \textcolor{red}{ANSWER}. Think very carefully about the task at hand, and analyze each part of speech in detail. If the specified part of speech does not exist in the sentence, strictly reply with NO. Otherwise, reply strictly with the erroneous sentence, marked with ANSWER, followed by Index: and the positions of the words you have changed. & "\textbf{Președinți} celor două Camere ale Parlamentului, Guvernul și Avocatul Poporului nu au comunicatul punctele lor de vedere asupra excepției de neconstituționalitate."

        Index: [0] \\
        \midrule
        \midrule
        
         Zero-shot & You are a grammar assistant tasked with introducing common unacceptable errors within a correct Romanian language sentence. For this task, \textcolor{green}{you will have to change the degree of an adjective erroneously} in the following sentence: \textcolor{red}{<<Aprobată prin ORDINUL nr. 304 din 19 octombrie 2020, publicat în Monitorul Oficial al României, Partea I, nr. 1026 din 4 noiembrie 2020.>>}. Think very carefully about the task at hand, and analyze each part of speech in detail. If the specified part of speech does not exist in the sentence, strictly reply with NO. Otherwise, reply strictly with the erroneous sentence, followed by Index: and the positions of the words you have changed. & "Aprobată prin ORDINUL nr. 304 din 19 octombrie 2020, publicat în Monitorul \textbf{cel mai Oficial} al României, Partea I, nr. 1026 din 4 noiembrie 2020." 
        
        Index: [14, 15, 16] \\
        \midrule
        Few-shot & You are a grammar assistant tasked with introducing common unacceptable errors within a correct Romanian language sentence. For this task, \textcolor{green}{you will have to change the degree of an adjective erroneously}. For example, the sentence \textcolor{blue}{<<Asta este soluția optimă.>>} will be turned into \textcolor{blue}{<<Asta este soluția cea mai optimă.>>}. For example, the sentence \textcolor{blue}{<<Gradul de colonel este superior celui de maior.>>} will be turned into \textcolor{blue}{<<Gradul de colonel este mai superior celui de maior.>>}. So, the sentence: \textcolor{red}{<<Aprobată prin ORDINUL nr. 304 din 19 octombrie 2020, publicat în Monitorul Oficial al României, Partea I, nr. 1026 din 4 noiembrie 2020.>>} will be turned into \textcolor{red}{ANSWER}. Think very carefully about the task at hand, and analyze each part of speech in detail. If the specified part of speech does not exist in the sentence, strictly reply with NO. Otherwise, reply strictly with the erroneous sentence, marked with ANSWER, followed by Index: and the positions of the words you have changed. & "Aprobată prin ORDINUL nr. 304 din 19 octombrie 2020, publicat în Monitorul \textbf{mai Oficial} al României, Partea I, nr. 1026 din 4 noiembrie 2020." 

        Index: [14, 15] \\
        \bottomrule
    \end{tabular}
    \caption{Zero-shot versus few-shot prompting output comparison. The green text represents the specific error, the red text is the clean sentence, and the blue text represents the common widespread Romanian errors used for two-shot prompting, adapted from the works of \citet{texbook1} and \citet{texbook2}.}
    \label{tab:zerovsfew}
\end{table}

The results of the manual LLM output analysis for the two-shot prompting technique have not revealed any significant inadequacies in the text corruption process. Thus, the disadvantage of using the two-shot prompting technique stems from a purely resource-based perspective, namely, the longer duration between the prompt submission and output extraction phases, as evidenced by objective time measurements conducted during the experiment. Duration assessment reveals that two-shot prompting is, on average, 11.37\% slower than zero-shot prompting, a difference that appears negligible within the confines of the 12-sample experiment, but becomes significant when creating a large synthetic dataset such as the RoLegalGEC. 

However, the main advantage of two-shot prompting, which greatly compensates for resource constraints, lies in the increased level of adequacy and correctness of the text corruption process, as shown by the most representative experimental samples in Table~\ref{tab:zerovsfew}, thereby improving the accuracy of the GED annotation process. 

For example, in the first shown sample, where we aim to create a subject-verb disagreement (VERB:SVA error type), the zero-shot output correctly identifies the subject and acknowledges that it should create a disagreement, but ultimately fails at detecting the verb in the sentence, and decides to alter the word next to the subject, which proves to be an attribute adjective, thus mistakenly tagging an adjective error as VERB:SVA. The added context provided by the two-shot prompting method appears to mitigate this issue by correctly identifying the verb and introducing a VERB:SVA error in the sentence.

The second example shows an attempt to change a noun inflection in the sentence. The zero-shot prompting confuses the verb "comunicat" (eng. "communicated") with a noun with the same spelling, "comunicat" (eng. "statement"), and corrupts the word as if it were a noun, compromising the annotation process by tagging a verb error as NOUN:INFL. On the contrary, the common Romanian mistake examples provided in the two-shot prompt steer the GPT-4o LLM's attention toward an actual noun and corrupt the token accordingly.

However, the third presented sample highlights why some error types are exceptions that we have decided to create using zero-shot prompting.  For this last sample, in which we vary an adjective degree, we see both prompting techniques producing outputs that, although different, are both adequate and correct corruptions of the clean corpus sentence. The sample results, paired with the fact that, for the two highlighted errors (NOUN:POSS and ADJ:FORM), there are very few possible word forms that shouldn't require excessive resources to conceptualize and eventually produce, have led to the decision to employ the simpler option, the zero-shot prompting technique, for these types of errors. 

To conclude RQ1, the two-shot prompting technique exhibits a notable disadvantage in resource utilization. In contrast, its advantage factually lies in the more accurate text corruption and annotation processes it enables. 

\subsection{Response for RQ2}

Intuitively, the inclusion of highly relevant data within a system's inner workings should correlate with a better understanding of the problem and, implicitly, higher system performance. However, in our case, the impact of augmenting the GEC input with GED error tags, thereby solving the correction task in a GEC-D setting rather than a classic GEC, seems to heavily depend on the architecture's ability to perceive and interpret the initial input data.

\begin{table} [!htpb]
    \centering
    \scriptsize
    \begin{tabular}{p{0.15\textwidth}p{0.4\textwidth}p{0.4\textwidth}}
\toprule
\textbf{Experimental Setup} & \textbf{Output for classic GEC task} & \textbf{Output for tag-augmented GEC-D task} \\
\midrule
RoBART-base + top-p & (2) Ordonatorii \textcolor{red}{principali principali} \textcolor{green}{de} credite cu \textcolor{red}{Rolurile} de Autoritate de management pentru programele operaționale, care beneficiază de prevederile alin. (1), sunt cei prevăzuți \textcolor{green}{în Anexă}. &  (2) Ordonatorii principali \textcolor{red}{pentru} credite cu \textcolor{red}{roluri} de Autoritate de management pentru programele operaționale, care beneficiază de prevederile alin. (1), sunt cei prevăzuți \textcolor{green}{în anexă}. \\
\midrule
RoBART-large + top-p & (2) Ordonatorii principali \textcolor{red}{printre} credite cu \textcolor{red}{Rolurile} de Autoritate de management pentru programele operaționale, care beneficiază de prevederile alin. (1), sunt cei prevăzuți \textcolor{green}{în anexă.} & (2) Ordonatorii principali \textcolor{red}{printre} credite cu \textcolor{red}{Rolurile} de Autoritate de management pentru programele operaționale, care beneficiază de prevederile alin. (1), sunt cei prevăzuți \textcolor{red}{înanexă.}  \\
\midrule
mBART + top-p & (2) Ordonatorii principali \textcolor{red}{la} credite cu \textcolor{red}{Rolurile} de Autoritate de management pentru programele operaționale, care beneficiază de prevederile alin. (1), sunt cei prevăzuți \textcolor{green}{în anexă}. & (2) Ordonatorii principali \textcolor{green}{de} credite cu \textcolor{red}{roluri} de Autoritate de management pentru programele operaționale, care beneficiază de prevederile alin. (1), sunt cei prevăzuți \textcolor{red}{înanexă}. \\
\midrule
\midrule
RoT5-base + top-p & (2) Ordonatorii principali \textcolor{green}{de} credite cu \textcolor{red}{roluri} de Autoritate de management pentru programele operaționale, care beneficiază de prevederile alin. (1), sunt cei prevăzuți \textcolor{green}{în anexă}. & (2) Ordonatorii principali \textcolor{green}{de} credite cu \textcolor{green}{rol} de Autoritate de management pentru programele operaționale, care beneficiază de prevederile alin. (1), sunt cei prevăzuți \textcolor{green}{în anexă}. \\
\midrule
mT5 + top-p & (2) Ordonatorii principali \textcolor{green}{de} credite cu \textcolor{green}{rol} de Autoritate de management pentru programele operaționale, care beneficiază de prevederile alin. (1), sunt cei prevăzuți \textcolor{red}{in annex}. & (2) Ordonatorii principali \textcolor{green}{de} credite cu \textcolor{green}{rol} de Autoritate de management pentru programele operaționale, care beneficiază de prevederile alin. (1), sunt cei prevăzuți \textcolor{green}{în anexă}. \\
\midrule
\midrule
RoBART-base + beam & (2) Ordonatorii principali \textcolor{green}{de} credite cu \textcolor{green}{Rol} de Autoritate de management pentru programele operaționale, care beneficiază de prevederile alin. (1) sunt cei prevăzuți în \textcolor{red}{anexaxă.} & (2) Ordonatorii principali \textcolor{green}{de} credite cu \textcolor{green}{rol} de Autoritate de management pentru programele operaționale, care beneficiază de prevederile alin. (1), sunt cei prevăzuți \textcolor{green}{în anexă}. \\
\midrule
RoBART-large + beam & (2) Ordonatorii principali \textcolor{green}{de} credite cu \textcolor{green}{rol} de Autoritate de management pentru programele operaționale, care beneficiază de prevederile \textcolor{red}{Alin} (1), sunt cei prevăzuți \textcolor{red}{înanexă}. & (2) Ordonatorii principali \textcolor{green}{de} credite cu \textcolor{red}{rolurile} de Autoritate de management pentru programele operaționale, care beneficiază de prevederile alin. (1), sunt cei prevăzuți \textcolor{green}{în anexă}. \\
\midrule
\midrule
RoT5-base + beam & (2) Ordonatorii principali \textcolor{green}{de} credite cu \textcolor{green}{rol} de Autoritate de management pentru programele operaționale, care beneficiază de prevederile alin. (1), sunt cei prevăzuți \textcolor{green}{în anexă}. & (2) Ordonatorii principali \textcolor{green}{de} credite cu \textcolor{green}{rol} de Autoritate de management pentru programele operaționale, care beneficiază de prevederile alin. (1), sunt cei prevăzuți \textcolor{green}{în anexă}. \\
\bottomrule
\end{tabular}
\caption{Comparison of inference results for different model setups between the classic GEC task setting and the tag-augmented GEC-D task setting, on the following corrupted source sentence: \textit{"(2) Ordonatorii principali \textcolor{red}{printre} credite cu \textcolor{red}{Rolurile} de Autoritate de management pentru programele operaționale, care beneficiază de prevederile alin. (1), sunt cei prevăzuți \textcolor{red}{înanexă.}"}. Erroneous sequences are marked in red, with their correct counterparts in green.}
\label{tab:rq2inference}
\end{table}

As concluded in the GEC results section, Table~\ref{tab:gecmetrics} paints a picture of the relative lack of efficiency of the BART architecture when using a top-p decoding strategy compared to setups employing the T5 architecture or the beam search strategy, which, as confirmed by the inference results presented in the first three rows of Table~\ref{tab:rq2inference}, leads to questionable choices and behavior in the correction of Romanian text. While the expectation is that aiding the training process with precise error tags should unequivocally optimize the performance of all types of models, the experiment results show that training in a GEC-D setting is unable to rectify the erratic training behavior exhibited by low-performing setups such as BART with a top-p strategy, further confusing the training process and reducing the overall score of \(F_{0.5}\) by a significant 0.032 for the RoBART-base variant, by 0.024 for the RoBART-large variant, and by 0.027 in a multi-lingual mBART setting.

In contrast, we analyze the better-performing setups, such as the T5 model with a top-p strategy or the BART architecture paired with a beam search strategy, that demonstrate a stronger ability to interpret the given input and solve the correction task. These setups substantially benefit from the inclusion of GED error tags, raising the overall \(F_{0.5}\) score by an average of 0.015 for all different variants of the aforementioned setups. 

The improvements are validated by the inference examples in Table~\ref{tab:rq2inference}, where the beam search RoBART-large setup provides a better, although not perfect, correction of the text when provided with input error tags in the GEC-D setup, by not inserting a new error in the text, as opposed to the classic GEC environment inference result; as for the three setups above it in the inference results table, all of them experience a slight correction struggle in a GEC setting, but perfectly correct the given text if provided with the GEC-D input error tags.  

Finally, the GEC-D training seems to have a modest impact upon the best-performing setups, namely the beam search T5 setups, given that for all variants of the T5 model paired with the beam search strategy, including the inference example provided in the last row of Table~\ref{tab:rq2inference}, both the GEC and GEC-D setups achieve the target correction. However, the slight impact is enough for the RoT5-base model, paired with a beam search strategy in the GEC-D setting, to be considered the overall best-performing correction model we have experimented with on the RoLegalGEC dataset.

To conclude RQ2, training a model in the GEC-D setting has a negative impact on weak architecture-strategy setups, a significant positive impact on strong correction setups, and a smoothed-out, neutral impact on optimized GEC setups, regardless of all other factors involved.

\subsection{Response for RQ3}

\paragraph{Language choice impact on Grammatical Error Detection} Since the mDistilBERT architecture has been pre-trained in multiple languages, the expectation is that the multilingual model would excel at detecting language-agnostic errors (MORPH, ORTH, PUNCT, SPELL, WO), when compared to the Romanian pre-trained variant, due to the additional training data in a context where the language of the data is least relevant. 

\begin{table} [htbp]
    \centering
    \scriptsize
    \begin{tabular}{p{0.2\textwidth}p{0.73\textwidth}}
\toprule
\textbf{Experimental Setup} & \textbf{Annotation Result} \\
\midrule
\midrule
RoDistilBERT inference & O PUNCT O O O O \textcolor{red}{NOUN} O O O O O CONJ O O O O O SPELL O MORPH O O O O O O O O O O O \textcolor{red}{O} \\
\midrule
mDistilBERT inference & O PUNCT O O O O NOUN:NUM O O O O O CONJ O O O O O SPELL O MORPH O O O O O O O O O O O PUNCT \\
\midrule
\midrule
Target detection & O PUNCT O O O O NOUN:NUM O O O O O CONJ O O O O O SPELL O MORPH O O O O O O O O O O O PUNCT \\
\bottomrule
\end{tabular}
\caption{Comparison of inference results between Romanian and multilingual variants of the DistilBERT-based error detection model, on the following corrupted source sentence: \textit{"8 \textcolor{red}{;} indicatorii chimici-cheie care pot determina \textcolor{red}{transformare} legate de resursele de apă \textcolor{red}{încă} de alte resurse naturale , \textcolor{red}{crae} pot \textcolor{red}{determinat} modificarea funcțiilor ecologice ale unei arii naturale protejate de interes comunitar \textcolor{red}{;}"}. Erroneous text sequences and incorrect annotations are marked in red.}
\label{tab:rq3gedlanguageinference}
\end{table}

The error-wise performance metrics in Table~\ref{tab:gedmetrics} confirm this hypothesis, with the mDistilBERT vastly improving the detection of morphology punctuation and word order errors, scoring the best out of all models in these categories, while also bringing slight improvements to orthography and spelling error detection performance when directly compared to the RoDistilBERT. The improvement can be visualized in Table~\ref{tab:rq3gedlanguageinference}, the mDistilBERT being able to detect and correctly annotate all language-agnostic errors, while the RoDistilBERT has not interpreted the ending semicolon of the sentence as an erroneous element.

The improvements brought by the multilingual DistilBERT version are slightly less evident for errors generated by the confusion list method. The mDistilBERT still produces the best results for detecting determiner and pronoun errors, but its performance starts to stagnate relative to the RoDistilBERT for preposition errors and even begins to drop for conjunction error detection.

Lastly, comparing the two DistilBERT models for part-of-speech errors generated by LLM pipelines yields mixed results. The mDistilBERT usage determined improvements for some of the part of speech characteristic detections, most notable being VERB:FORM, NOUN:INFL and NOUN:NUM, the latter improvement being shown in Table~\ref{tab:rq3gedlanguageinference}, where the mDistilBERT noticed the singular noun "transformare" (en. "transformation") placed before a plural adjectival attribute "legate de" (en. "linked with") and correctly interprets the error as a NOUN:NUM error instead of a generic NOUN error, like the RoDistilBERT deduced. 

The downside of the mDistilBERT is that, while it brings slight improvements to a significant number of areas, the errors in which RoDistilBERT performs better highlight severe performance decreases and issues for the multilingual version, evidenced in errors such as the generic NOUN and ADJ errors, indicating a level of inconsistency in the results determined by the multiple language training. To conclude the language choice comparison for GED experiments on the RoLegalGEC dataset, the Romanian pre-trained model exhibits relatively weak metric performance, but the results are stable and consistent in most detection areas; the usage of multilingual models undeniably improves model performance and ultimately leads to better inferences, but the trade-off consists of significant erratic performance gaps between different sectors of our error taxonomy.

\paragraph{Language choice impact on Grammatical Error Correction}
In the case of error correction experiments for the GEC and GEC-D tasks on the RoLegalGEC dataset, through analysis of the impact of pre-training language choice on the performance metrics shown in Table~\ref{tab:gecmetrics}, we can observe that a multilingual model variants bring significant spikes to the precision metric in all available experiment setups, indicating the fact that these models have a higher capacity to identify initially correct text and refrain from needlessly and erroneously "correct" it. The expected downside of multilingual variant usage, however, is a relative inability to correct existing erroneous sequences compared with models trained solely on Romanian data, resulting in low recall.

\begin{table} [th]
    \centering
    \scriptsize
    \begin{tabular}{p{0.15\textwidth}p{0.4\textwidth}p{0.4\textwidth}}
\toprule
\textbf{Experimental Setup} & \textbf{Base Romanian model variant output} & \textbf{Multilingual model variant output} \\
\midrule
\midrule
BART + top-p + GEC & (2) Ordonatorii \textcolor{red}{principali principali} \textcolor{green}{de} credite cu \textcolor{red}{Rolurile} de Autoritate de management pentru programele operaționale, care beneficiază de prevederile alin. (1), sunt cei prevăzuți \textcolor{green}{în Anexă}. & (2) Ordonatorii principali \textcolor{red}{la} credite cu \textcolor{red}{Rolurile} de Autoritate de management pentru programele operaționale, care beneficiază de prevederile alin. (1), sunt cei prevăzuți \textcolor{green}{în anexă}. \\
\midrule
T5 + top-p + GEC & (2) Ordonatorii principali \textcolor{green}{de} credite cu \textcolor{red}{roluri} de Autoritate de management pentru programele operaționale, care beneficiază de prevederile alin. (1), sunt cei prevăzuți \textcolor{green}{în anexă}. & (2) Ordonatorii principali \textcolor{green}{de} credite cu \textcolor{green}{rol} de Autoritate de management pentru programele operaționale, care beneficiază de prevederile alin. (1), sunt cei prevăzuți \textcolor{red}{in annex}. \\
\midrule
BART + top-p + GEC-D & (2) Ordonatorii principali \textcolor{red}{pentru} credite cu \textcolor{red}{roluri} de Autoritate de management pentru programele operaționale, care beneficiază de prevederile alin. (1), sunt cei prevăzuți \textcolor{green}{în anexă}. & (2) Ordonatorii principali \textcolor{green}{de} credite cu \textcolor{red}{roluri} de Autoritate de management pentru programele operaționale, care beneficiază de prevederile alin. (1), sunt cei prevăzuți \textcolor{red}{înanexă}. \\
\midrule
BART + beam + GEC-D & (2) Ordonatorii principali \textcolor{green}{de} credite cu \textcolor{green}{rol} de Autoritate de management pentru programele operaționale, care beneficiază de prevederile alin. (1), sunt cei prevăzuți \textcolor{green}{în anexă}. & (2) Ordonatorii principali \textcolor{red}{printre} credite cu \textcolor{green}{rol} de Autoritate de management pentru programele operaționale, care beneficiază de prevederile alin. (1), sunt cei prevăzuți \textcolor{green}{în anexă}. \\
\midrule
BART + beam + GEC & (2) Ordonatorii principali \textcolor{green}{de} credite cu \textcolor{green}{Rol} de Autoritate de management pentru programele operaționale, care beneficiază de prevederile alin. (1) sunt cei prevăzuți în \textcolor{red}{anexaxă.} & (2) Ordonatorii principali \textcolor{green}{de} credite cu \textcolor{green}{rol} de Autoritate de management pentru programele operaționale, care beneficiază de prevederile alin. (1), sunt cei prevăzuți \textcolor{green}{în anexă}. \\
\midrule
T5 + beam + GEC-D & (2) Ordonatorii principali \textcolor{green}{de} credite cu \textcolor{green}{rol} de Autoritate de management pentru programele operaționale, care beneficiază de prevederile alin. (1), sunt cei prevăzuți \textcolor{green}{în anexă}. & (2) Ordonatorii principali \textcolor{green}{de} credite cu \textcolor{green}{rol} de Autoritate de management pentru programele operaționale, care beneficiază de prevederile alin. (1), sunt cei prevăzuți \textcolor{green}{în anexă}. \\
\bottomrule
\end{tabular}
\caption{Comparison of inference results between Romanian and multilingual variants of the error correction models, on the following corrupted source sentence: \textit{"(2) Ordonatorii principali \textcolor{red}{printre} credite cu \textcolor{red}{Rolurile} de Autoritate de management pentru programele operaționale, care beneficiază de prevederile alin. (1), sunt cei prevăzuți \textcolor{red}{înanexă.}"}. Erroneous sequences are marked in red, with their correct counterparts in green.}
\label{tab:rq3geclanguageinference}
\end{table}

 The most insightful illustration of this phenomenon is shown in the first row of Table~\ref{tab:rq3geclanguageinference}, where we compare inference results between RoBART-base and mBART in the GEC top-p search setup. The Romanian variant manages to correct two of the three existing errors in the corrupted sentence, but mistakenly doubles a word and thus generates an additional error; meanwhile, the multilingual version refrains from generating errors, improving the precision in the process, but corrects fewer errors than the Romanian variant, consequently affecting the recall metric. 

Due to the highly improved precision stats, the improvements that the multilingual variants are supposed to bring over the Romanian variants for low-performing models seem significant metric-wise, looking at the overall \(F_{0.5}\) metric, backed by inference examples such as the GEC BART beam search setup, where the mBART perfectly corrects a sentence that the RoBART-base struggled with); however, these improvements might be difficult to identify through inference results, due to trade-offs between false positives and false negatives in identifying and correcting errors, illustrated in Table~\ref{tab:rq3geclanguageinference} for the setups that employ the top-p decoding strategy.

Regarding the top-performing models in Romanian pre-trained settings, the multilingual variant scores similar, if not the same, \(F_{0.5}\) scores as the Romanian counterparts. But because of the sheer precision-recall imbalance between multilingual and Romanian models, we conclude that, even though the multilingual models' metric performance is comparable, if not even better, the Romanian pre-trained models represent the most secure and efficient solutions for text correction on the RoLegalGEC dataset. The experiments on the GEC-D BART beam search setup (where the RoBART corrects every error while the mBART misses one) and the GEC-D T5 beam search setup (where both variants achieve perfect corrections, but the context of performance metrics proves the instability of the multilingual variant) thoroughly support our conclusion.

\paragraph{Overall conclusion for language choice impact} The analysis of the impact of pre-training language choice across both GED and GEC environments have led to a similar conclusion, which is that the choice of Romanian pre-training model translates to a secure, stable and consistent experiment setup that, while it can reach performance peaks similar to its multilingual counterpart, generally it tends to perform relatively worse. In contrast, the choice of a multilingual pre-trained setup yields overall performance improvements and fixes glaring issues reported by the Romanian variants, but it tends to falter in other areas, leading to inconsistent results. In conclusion, the relevance of areas where the multilingual variant is less efficient should be the primary deciding factor when choosing the pre-training language; for the RoLegalGEC dataset training, this translates into choosing Romanian RoT5-base as the best correction model and the multilingual mDistilBERT as the best detection model.

\paragraph{Model size impact on Grammatical Error Detection} On a closer analysis of the Sequence Tagging performance metric results illustrated in Table~\ref{tab:gedmetrics}, we can observe that the usage of the larger, more complex variant of the BERT encoder provided by the SeqTag-large architecture significantly improves the annotation ability of the overall architecture for errors that are infrequent, but important for Romanian, even managing to fix the glaring performance issues of the SeqTag-base architecture (e.g. raising the PRON \(F_{0.5}\) score from 0.354 to 0.711, the PREP score from 0.131 to 0.493, and the ADJ score from 0.404 to 0.546); however, there is little, if any, difference registered between the variants when it comes to the more frequent, mainly language-agnostic errors (MORPH, ORTH, PUNCT, SPELL, WO), highlighting the saturated potential of the model and confirming that the detection of unspecialized errors can be perfected even by using smaller, more compact architectures, such as the SeqTag-base.

The aforementioned facts result in a slight overall performance increase from the SeqTag-base to the SeqTag-large model, exemplified by the inference results in Table~\ref{tab:rq3gedsizeinference}, where the SeqTag-large inference, while not perfect like the mDistilBERT inference, manages to correct some of the errors missed by the SeqTag-base; the performance increase, however, is backed by massive improvements in the areas most relevant to the RoLegalGEC dataset research.   

\begin{table} [htbp]
    \centering
    \scriptsize
    \begin{tabular}{p{0.2\textwidth}p{0.73\textwidth}}
\toprule
\textbf{Experimental Setup} & \textbf{Annotation Result} \\
\midrule
\midrule
SeqTag-base inference & O \textcolor{red}{O} O O O O NOUN:NUM O O O O O CONJ O O O O O SPELL O \textcolor{red}{O} O O O O O O O O O O O PUNCT \\
\midrule
SeqTag-large inference & O PUNCT O O O O NOUN:NUM O O O O O CONJ O O O O O SPELL O \textcolor{red}{O} O O O O O O O O O O O PUNCT \\
\midrule
\midrule
Target detection & O PUNCT O O O O NOUN:NUM O O O O O CONJ O O O O O SPELL O MORPH O O O O O O O O O O O PUNCT \\
\bottomrule
\end{tabular}
\caption{Comparison of inference results between the base and large size variants of the Sequence Tagging error detection model, on the following corrupted source sentence: \textit{"8 \textcolor{red}{;} indicatorii chimici-cheie care pot determina \textcolor{red}{transformare} legate de resursele de apă \textcolor{red}{încă} de alte resurse naturale , \textcolor{red}{crae} pot \textcolor{red}{determinat} modificarea funcțiilor ecologice ale unei arii naturale protejate de interes comunitar \textcolor{red}{;}"}. Erroneous text sequences and incorrect annotations are marked in red.}
\label{tab:rq3gedsizeinference}
\end{table}

\paragraph{Model size impact on Grammatical Error Correction}

The results of testing both base and large variants of GEC architectures fine-tuned on the RoLegalGEC dataset, and the subsequent comparison of the relevant performance metrics in Table~\ref{tab:gecmetrics}, highlight a clear pattern that defines the impact of a larger, more resource-intensive architecture on the ability to correct Romanian legal data. Concretely, the usage of the RoBART-large model has an all-around negative impact on correction performance, regardless of specific task or decoding strategy, with the $F_{0.5}$ score plummeting by an average of 0.007 compared to RoBART-base experiments, a decline depicted by relevant inference examples in Table~\ref{tab:rq3gecsizeinference}, where the RoBART-large variant output creates an additional error in a top-p GEC setting, and manages to not accurately correct a sentence that the RoBART-base model handled perfectly in the beam search GEC-D setting.

\begin{table} [th]
    \centering
    \scriptsize
    \begin{tabular}{p{0.15\textwidth}p{0.4\textwidth}p{0.4\textwidth}}
\toprule
\textbf{Experimental Setup} & \textbf{Base Romanian model variant output} & \textbf{Large Romanian model variant output} \\
\midrule
\midrule
BART + top-p + GEC &  În cazul candidatului independent se înscrie \textcolor{green}{și} funcția pentru care acesta a \textcolor{red}{candidați} \textcolor{green}{:} primar sau consilier \textcolor{green}{local}. & În cazul candidatului independent se înscrie \textcolor{green}{și} funcția pentru care 
\textcolor{red}{aceșția} a \textcolor{red}{candidați} \textcolor{green}{:} primar sau consilier \textcolor{green}{local}. \\
\midrule
BART + beam + GEC-D & În cazul candidatului independent se înscrie \textcolor{green}{și} funcția pentru care acesta a \textcolor{green}{candidat} \textcolor{green}{:} primar sau consilier \textcolor{green}{local}. & În cazul candidatului independent se înscrie \textcolor{green}{și} funcția pentru care acesta a \textcolor{green}{candidat} \textcolor{green}{:} primar sau consilier \textcolor{red}{locală}. \\
\midrule
\midrule
T5 + top-p + GEC & În cazul candidatului independent se înscrie \textcolor{red}{cu} funcția pentru care acesta a \textcolor{green}{candidat} \textcolor{red}{,} primar sau consilier \textcolor{green}{local}. & În cazul candidatului independent se înscrie \textcolor{green}{și} funcția pentru care acesta a \textcolor{green}{candidat} \textcolor{green}{:} primar sau consilier \textcolor{green}{local}. \\
\midrule
T5 + top-p + GEC-D & În cazul candidatului independent se înscrie \textcolor{green}{și} funcția pentru care acesta a \textcolor{red}{candidează} \textcolor{green}{:} primar sau consilier \textcolor{green}{local}. & În cazul candidatului independent se înscrie \textcolor{green}{și} funcția pentru care acesta a \textcolor{green}{candidat} \textcolor{green}{:} primar sau consilier \textcolor{green}{local}.\\
\midrule
T5 + beam + GEC & În cazul candidatului independent se înscrie \textcolor{green}{și} funcția pentru care acesta a \textcolor{green}{candidat} \textcolor{green}{:} primar sau consilier \textcolor{green}{local}. & În cazul candidatului independent se înscrie \textcolor{green}{și} funcția pentru care acesta a \textcolor{green}{candidat} \textcolor{green}{:} primar sau consilier \textcolor{green}{local}. \\
\bottomrule
\end{tabular}
\caption{Comparison of inference results between Romanian base and large variants of the error correction models, on the following corrupted source sentence: \textit{"În cazul candidatului independent se înscrie \textcolor{red}{sau} funcția pentru care acesta a \textcolor{red}{candidați} \textcolor{red}{.} primar sau consilier \textcolor{red}{locală}."}. Erroneous sequences are marked in red, with their correct counterparts in green.}
\label{tab:rq3gecsizeinference}
\end{table}

In contrast to BART experiments, the RoT5-large model has either a neutral or positive impact on correction performance compared to RoT5-base model results in the same experimental setup. The large variant improves $F_{0.5}$ scores by an average of 0.01 and, when looking at the inference results in Table~\ref{tab:rq3gecsizeinference}, the RoT5-large outputs accurate corrections in all relevant setup instances, improving upon slightly erroneous attempts of RoT5-base inference.

\paragraph{Overall conclusions for model size impact}

Unlike the analysis of the impact of language choice, which had a complex resolution that depended on a series of factors and variables, the conclusion regarding the impact of model size on metric performance across both detection and correction tasks appears straightforward and strictly boils down to architecture selection. The choice of a larger architecture significantly improves the performance of the BERT encoder embedded within the GED Sequence Tagger, as well as the correction T5 architecture, but consistently harms the performance of an architecture such as the BART when fine-tuned and tested on the Romanian RoLegalGEC parallel legal dataset.   

\subsection{Limitations}
A significant limitation in pursuing a realistic synthetic dataset, as the RoLegalGEC concept proposes, is that, due to the intricacy of Romanian, it is difficult to devise a compact set of simple grammatical rules that covers most cases models might encounter. While there are a few straightforward rules, most of them regarding character-level errors, that we have made sure to include as options in our work, a clear and comprehensive grammatical guideline would certainly simplify the corruption procedure and create a more realistic synthetic dataset.

Another limitation relates to the computational intensity of LLM-based generation due to the prompt complexity required to generate correct and reliable text corruption, a phenomenon more egregious when considering the few-shot prompting technique, which is why we use the LLM technique sparingly, reserving it for high complexity errors. For the other corruption techniques, the only limitation arises when certain controlled substitutions require communication with online dictionaries, which consume substantially more resources than simpler techniques; however, such cases are rare.

Additionally, a conceptual limitation concerns the ability to assess the impact of a shift in experimental methodology, such as architecture, setup, or generation strategy, on the model's overall performance and on each type of error separately. The extent of our experiments has highlighted significant correlations between experimental methodology and model performance, including shifts across several error types. However, the same results exhibit unexpected or seemingly random behavior for other taxonomy elements, which would require considerable resources and computationally intensive training to determine whether they constitute unforeseen correlations or pure outliers, thus establishing the full explainability of the detection and correction procedures. 

\section{Conclusions and Future Work}

In this paper, we have developed RoLegalGEC, the first Romanian synthetic parallel dataset for detecting and correcting grammatical errors in the legal domain. It consists of 350,000 samples acquired from extensive, expert-sourced corpora. The dataset includes a variety of human-like erroneous utterances specific to Romanian, in accordance with our proposed 20-error taxonomy for Romanian, which intends to cover most basic language needs. 
Along with the dataset, we have established several synthetic data generation procedures, meant to create realistic erroneous sequences from correct examples. The methods ranged from classic, intuitive, and simple mechanisms to more innovative experimental strategies, supported by grammatical expertise in the form of our proposed common Romanian error database, validated by specialized literature, an overall set of data generation procedures that we have devised to successfully consider and accommodate each error type in the taxonomy, thus providing complete control over desired dataset statistics.

Additionally, using the novel dataset, we have evaluated several models for both detecting and correcting grammatical errors. We have conducted an in-depth error-wise analysis of detection model results, confirming that pre-trained models capture the essence of the specific language better, while basic architectures master the fundamentals of text construction. For the correction module, our proposed experiments on the niche dataset have highlighted the need for an adaptable text-to-text architecture, supported by a complex selection strategy, to generate the most accurate text corrections. Finally, we have formally defined an additional correction task, GEC-D, which is useful for users who value the analysis provided by the GED task and wish to leverage it to improve the correction quality. We have provided models intended to solve the GEC-D task by improving our GEC architectures using GED error tags. Although detrimental to low-performing GEC setups, they represent an overall improvement for GEC models that can initially comprehend and correct text.

In future research, we aim to further streamline the error generation process, for example, by addressing the identified limitations of the text-corruption and model-testing pipelines. Future objectives of this research also include a rigorous process of verifying and optimizing the RoLegalGEC dataset and task formulations, with the objective of integrating our novel legal dataset into an established Romanian NLP benchmark, namely LiRo \citep{liro2021}. The LiRo benchmark facilitates progress monitoring for models trained on their provided tasks and datasets, and integrating our novel dataset would align with our goal of strengthening the Romanian NLP research.  

\bibliographystyle{rusnat}  
\bibliography{references}  
\end{document}